# Conflict-Directed Backjumping Revisited

**Xinguang Chen** XINGUANG@CS.UALBERTA.CA
*Department of Computing Science, University of Alberta*
*Edmonton, Alberta, Canada T6G 2H1*

**Peter van Beek** VANBEEK@UWATERLOO.CA
*Department of Computer Science, University of Waterloo*
*Waterloo, Ontario, Canada N2L 3G1*

## Abstract

In recent years, many improvements to backtracking algorithms for solving constraint satisfaction problems have been proposed. The techniques for improving backtracking algorithms can be conveniently classified as *look-ahead* schemes and *look-back* schemes. Unfortunately, look-ahead and look-back schemes are not entirely orthogonal as it has been observed empirically that the enhancement of look-ahead techniques is sometimes counterproductive to the effects of look-back techniques. In this paper, we focus on the relationship between the two most important look-ahead techniques—using a variable ordering heuristic and maintaining a level of local consistency during the backtracking search—and the look-back technique of conflict-directed backjumping (CBJ). We show that there exists a "perfect" dynamic variable ordering such that CBJ becomes redundant. We also show theoretically that as the level of local consistency that is maintained in the backtracking search is increased, the less that backjumping will be an improvement. Our theoretical results partially explain why a backtracking algorithm doing more in the look-ahead phase cannot benefit more from the backjumping look-back scheme. Finally, we show empirically that adding CBJ to a backtracking algorithm that maintains generalized arc consistency (GAC), an algorithm that we refer to as GAC-CBJ, can still provide orders of magnitude speedups. Our empirical results contrast with Bessière and Régin's conclusion (1996) that CBJ is useless to an algorithm that maintains arc consistency.

## 1. Introduction

Constraint satisfaction problems (CSPs) are a generic problem solving framework. A constraint satisfaction problem consists of a set of variables, each associated with a domain of values, and a set of constraints. Each of the constraints is expressed as a relation, defined on some subset of the variables, denoting the consistent value assignments that satisfy the constraint. A solution to a CSP is an assignment of a value to every variable, in such a way that every constraint is satisfied.

Constraint satisfaction problems are usually solved by search methods, among which the backtracking algorithm and its improvements are widely used. The techniques for improving backtracking algorithms can be conveniently classified as *look-ahead schemes* and *look-back schemes* (Dechter, 1992). Look-ahead schemes are invoked whenever the algorithm is preparing to extend the current partial solution. Look-ahead schemes include the functions that choose the next variable to be instantiated, choose the next value to give to the current variable, and reduce the search space by maintaining a certain level of local consistency during the search (e.g., Bacchus & van Run, 1995; Bessière & Régin, 1996;

Haralick & Elliott, 1980; McGregor, 1979; Nadel, 1989; Sabin & Freuder, 1994). Look-back schemes are invoked whenever the algorithm encounters a dead-end and prepares for the backtracking step. Look-back schemes include the functions that decide how far to backtrack by analyzing the reasons for the dead-end (backjumping) and decide what new constraints to record so that the same conflicts do not arise again later in the search (e.g., Bruynooghe, 1981; Dechter, 1990; Frost & Dechter, 1994; Gaschnig, 1978; Prosser, 1993b; Schiex & Verfaillie, 1994).

A backtracking algorithm can be a hybrid of both look-ahead and look-back schemes (Prosser, 1993b). In this paper, we focus on the relationship between the two most important look-ahead techniques—using a variable ordering heuristic and maintaining a level of local consistency during the backtracking search—and the look-back technique of conflict-directed backjumping (CBJ) (Prosser, 1993b). Unfortunately, these look-ahead and look-back schemes are not entirely orthogonal as it can be observed in previous experimental work that as the level of consistency that is maintained in the backtracking search is increased and as the variable ordering heuristic is improved, the effects of CBJ are diminished (Bacchus & van Run, 1995; Bessière & Régin, 1996; Prosser, 1993a, 1993b). For example, it can be observed in Prosser's (1993b) experiments that, given a static variable ordering, increasing the level of local consistency maintained from none to the level of forward checking, diminishes the effects of CBJ. Bacchus and van Run (1995) observe from their experiments that adding a dynamic variable ordering (an improvement over a static variable ordering) to a forward checking algorithm diminishes the effects of CBJ. In their experiments the effects are so diminished as to be almost negligible and they present an argument for why this might hold in general. Bessière and Régin (1996) observe from their experiments that simultaneously increasing the level of local consistency even further to arc consistency and further improving the dynamic variable ordering heuristic diminishes the effects of CBJ so much that, in their implementation, the overhead of maintaining the data structures for backjumping actually slows down the algorithm. They conjecture that when arc consistency is maintained and a good variable ordering heuristic is used, "CBJ becomes useless".

In this paper, we present theoretical results that deepen our understanding of the relationship between look-ahead techniques and the CBJ look-back technique. We show that there exists a "perfect" dynamic variable ordering for the chronological backtracking algorithm such that CBJ becomes redundant. The more that a variable ordering heuristic is consistent with the "perfect" heuristic, the less chance CBJ has to reduce the search effort. We also show that CBJ and an algorithm that maintains strong $k$-consistency in the backtracking search are incomparable in that each can be exponentially better than the other. This result is refined by introducing the concept of *backjump level* in the execution of a backjumping algorithm and showing that an algorithm that maintains strong $k$-consistency never visits more nodes than a backjumping algorithm that is allowed to backjump at most $k$ levels. Thus, as the level of local consistency that is maintained in the backtracking search is increased, the less that backjumping will be an improvement. Together, our theoretical results partially explain why a backtracking algorithm doing more in the look-ahead phase cannot benefit more from the backjumping look-back scheme. Our results also extend the partial ordering of backtracking algorithms presented by Kondrak and van Beek (1997) to include backtracking algorithms and their CBJ hybrids that maintain levels of local con-sistency beyond forward checking, including the important algorithms that maintain arc consistency.

We also present empirical results that show that, although the effects of CBJ may be diminished, adding CBJ to a backtracking algorithm that maintains generalized arc consistency (GAC), an algorithm that we refer to as GAC-CBJ, can still provide orders of magnitude speedups. Our empirical results contrast with Bessière and Régin's (1996) conclusion that CBJ is useless to an algorithm that maintains arc consistency.

## 2. Background

In this section, we formally define constraint satisfaction problems, and briefly review local consistency and the search tree explored by a backtracking algorithm.

### 2.1 Constraint Satisfaction Problems

**Definition 1 (CSP)** *An instance of a constraint satisfaction problem is a tuple* $P = (\mathcal{V}, \mathcal{D}, \mathcal{C})$, *where*[1]

- $\mathcal{V} = \{x_1, \ldots, x_n\}$ *is a finite set of* $n$ *variables,*
- $\mathcal{D} = \{dom(x_1), \ldots, dom(x_n)\}$ *is a set of domains. Each variable* $x \in \mathcal{V}$ *is associated with a finite domain of possible values,* $dom(x)$. *The maximum domain size* $max_{x\in\mathcal{V}}|dom(x)|$ *is denoted by* $d$,
- $\mathcal{C} = \{C_1, \ldots, C_m\}$ *is a finite set of* $m$ *constraints or relations. Each constraint* $C \in \mathcal{C}$ *is a pair* $(vars(C), rel(C))$, *where*
  - $vars(C) = \{x_{i_1}, \ldots, x_{i_{r_i}}\}$ *is an ordered subset of the variables, called the constraint scope or scheme, the size of* $vars(C)$ *is known as the arity of the constraint. If the arity of the constraint is equal to 2, it is called a binary constraint. A* non-binary *constraint is a constraint with arity greater than 2. The maximum arity of the constraints in* $\mathcal{C}$, $max_{C\in\mathcal{C}}|vars(C)|$, *is denoted by* $r$,
  - $rel(C)$ *is a subset of the Cartesian product* $dom(x_{i_1})\times\cdots\times dom(x_{i_{r_i}})$ *that specifies the allowed combinations of values for the variables in* $vars(C)$. *An element of the Cartesian product* $dom(x_{i_1}) \times \cdots \times dom(x_{i_{r_i}})$ *is called a tuple on* $vars(C)$. *Thus,* $rel(C)$ *is often regarded as a set of tuples over* $vars(C)$.

In the following, we assume that for any variable $x \in \mathcal{V}$, there is at least one constraint $C \in \mathcal{C}$ such that $x \in vars(C)$. By definition, a tuple over a set of variables $X = \{x_1, \ldots, x_k\}$ is an ordered list of values $(a_1, \ldots, a_k)$ such that $a_i \in dom(x_i)$, $i = 1, \ldots, k$. A tuple over $X$ can also be regarded as a set of variable-value pairs $\{x_1 \leftarrow a_1, \ldots, x_k \leftarrow a_k\}$. Furthermore, a tuple over $X$ can be viewed as a function $t : X \rightarrow \cup_{x\in X} dom(x)$ such that for each variable $x \in X$, $t[x] \in dom(x)$. For a subset of variables $X' \subseteq X$, we use $t[X']$ to denote a tuple over $X'$ by restricting $t$ over $X'$. We also use $vars(t)$ to denote the set of variables for tuple $t$.

1. Throughout the paper, we use $n$, $d$, $m$, and $r$ to denote the number of variables, the maximum domain size, the number of constraints, and the maximum arity of the constraints in the CSP, respectively.

An *assignment* to a set of variables $X$ is a tuple over $X$. We say an assignment $t$ to $X$ is *consistent* with a constraint $C$ if either $vars(C) \not\subseteq X$ or $t[vars(C)] \in rel(C)$. A *partial solution* to a CSP is an assignment to a subset of variables. We say a partial solution is *consistent* if it is consistent with each of the constraints. A *solution* to a CSP is a consistent partial solution over all the variables. If no solution exists, the CSP is said to be insoluble. A CSP is *empty* if either one of its variables has an empty domain or one of its constraints has an empty set of tuples. Obviously, an empty CSP is insoluble. Given two CSP instances $P_1$ and $P_2$, we say $P_1 = P_2$ if they have exactly the same set of variables, the same set of domains and the same set of constraints; i.e., they are syntactically the same.

**Definition 2 (projection)** *Given a constraint $C$ and a subset of variables $S \subseteq vars(C)$, the projection $\pi_S C$ is a constraint, where $vars(\pi_S C) = S$ and $rel(\pi_S C) = \{t[S] \mid t \in rel(C)\}$.*

**Definition 3 (selection)** *Given a constraint $C$ and an assignment $t$ to a subset of variables $X \subseteq vars(C)$, the selection $\sigma_t C$ is a constraint, where $vars(\sigma_t C) = vars(C)$ and $rel(\sigma_t C) = \{s \mid s[X] = t$ and $s \in rel(C)\}$.*

### 2.2 Local Consistency

An *inconsistency* is a consistent partial solution over some of the variables that cannot be extended to additional variables and so cannot be part of any global solution. If we are using a backtracking search to find a solution, such an inconsistency can lead to a dead end in the search. This insight has led to the definition of properties that characterize the level of consistency of a CSP and to the development of algorithms for achieving these levels of consistency by removing inconsistencies (e.g., Mackworth, 1977a; Montanari, 1974), and to effective backtracking algorithms for finding solutions to CSPs that maintain a level of consistency during the search (e.g., Gaschnig, 1978; Haralick & Elliott, 1980; McGregor, 1979; Sabin & Freuder, 1994).

Mackworth (1977a) defines three properties of binary CSPs that characterize local consistencies: node, arc, and path consistency. Mackworth (1977b) generalizes arc consistency to non-binary CSPs.

**Definition 4 (arc consistency)** *Given a constraint $C$ and a variable $x \in vars(C)$, a value $a \in dom(x)$ is* supported *in $C$ if there is a tuple $t \in rel(C)$, such that $t[x] = a$. $t$ is then called a* support *for $\{x \leftarrow a\}$ in $C$. $C$ is* arc consistent *if for each of the variables $x \in vars(C)$, and each of the values $a \in dom(x)$, $\{x \leftarrow a\}$ is supported in $C$. A CSP is* arc consistent *if each of its constraints is arc consistent.*

Freuder (1978) generalizes node, arc, and path consistency, to $k$-consistency.

**Definition 5 ($k$-consistency)** *A CSP is $k$-consistent if and only if given any consistent partial solution over $k-1$ distinct variables, there exists an instantiation of any $k^{th}$ variable such that the partial solution plus that instantiation is consistent. A CSP is strongly $k$-consistent if it is $j$-consistent for all $1 \leq j \leq k$.*

For binary CSPs, node, arc and path consistency correspond to one-, two- and three-consistency, respectively. However, the definition of $k$-consistency does not require the CSP to be binary and arc consistency is not the same as two-consistency for non-binary CSPs. A strongly $n$-consistent CSP has the property that any consistent partial solution can be successively extended to a full solution of the CSP without backtracking.

### 2.3 Search Tree and Backtracking Algorithms

The idea of a *backtracking algorithm* is to extend partial solutions. At each stage, an uninstantiated variable is selected and assigned a value from its domain to extend the current partial solution[2]. Constraints are used to check whether such an extension may lead to a possible solution of the CSP and to prune subtrees containing no solutions based on the current partial solution. During a backtracking search, the variables can be divided into three sets: *past variables* (already instantiated), *current variable* (now being instantiated), and *future variables* (not yet instantiated). A *dead-end* occurs when all values of the current variable are rejected as not leading to a full solution. In such a case, some instantiated variables become *uninstantiated*; i.e., they are removed from the current partial solution. This process is called *backtracking*. If only the most recently instantiated variable becomes uninstantiated then it is called *chronological backtracking*; otherwise, it is called *backjumping*. A backtracking algorithm terminates when all possible assignments have been tested or a certain number of solutions have been found.

A backtracking search may be seen as a *search tree* traversal. In this approach we identify tuples (assignments of values to variables) with nodes: the empty tuple is the root of the tree, the first level nodes are 1-tuples (representing an assignment of a value to a single variable), the second level nodes are 2-tuples, and so on. The levels closer to the root are called *shallower levels* and the levels farther from the root are called *deeper levels*. Similarly, the variables corresponding to these levels are called shallower and deeper. We say that a backtracking algorithm visits a node in the search tree if at some stage of the algorithm's execution the current partial solution identifies the node. The nodes visited by a backtracking algorithm form a subset of all the nodes belonging to the search tree. We call this subset, together with the connecting edges, the *backtrack tree* generated by a backtracking algorithm.

The backtracking algorithm conflict-directed backjumping (CBJ) (Prosser, 1993b) maintains a *conflict set* for every variable. Every time an instantiation of the current variable $x_i$ is in conflict with an instantiation of some past variable $x_h$, the variable $x_h$ is added to the conflict set of $x_i$. When there are no more values to be tried for the current variable $x_i$, CBJ backtracks to the deepest variable $x_h$ in the conflict set of $x_i$. At the same time, the variables in the conflict set of $x_i$, with the exception of $x_h$, are added to the conflict set of $x_h$, so that no information about conflicts is lost.

Throughout the paper we refer to the following backtracking algorithms (see Kondrak & van Beek, 1997; Prosser, 1993b for detailed explanations and examples of most of these algorithms): chronological backtracking (BT), backjumping (BJ) (Gaschnig, 1978), conflict-directed backjumping (CBJ) (Prosser, 1993b), forward checking (FC) (Haralick & Elliott, 1980; McGregor, 1979), forward checking and conflict-directed backjumping (FC-CBJ)

2. Throughout this paper, we assume that a static *value* ordering is used in the backtracking search.

(Prosser, 1993b), maintaining arc consistency (MAC) (Gaschnig, 1978; Sabin & Freuder, 1994), and maintaining arc consistency and conflicted-directed backjumping (MAC-CBJ) (Prosser, 1995).

## 3. Variable Ordering Heuristics and Backjumping

In this section, we present theoretical results that deepen our understanding of the relationship between the look-ahead technique of using a variable ordering heuristic and the look-back technique of CBJ.

In previous work, Kondrak and van Beek (1997) show that, given the same deterministic static or dynamic variable ordering heuristic, CBJ never visits more nodes than BT. Bacchus and van Run (1995) show that BJ, a restricted version of CBJ, visits exactly the same nodes as BT if the fail-first dynamic variable ordering heuristic is used. Previous empirical work shows that the number of nodes that CBJ saves depends on the variable ordering heuristic used (Bacchus & van Run, 1995; Bessière & Régin, 1996; Prosser, 1993b).

We show that, given a CSP and a variable ordering for CBJ, there exists a "perfect" variable ordering for the chronological backtracking algorithm (BT) such that BT never visits more nodes than CBJ. The more that a variable ordering heuristic is consistent with the "perfect" heuristic, the less chance CBJ has to reduce the search effort.

We first consider the case of insoluble CSPs. When CBJ is applied to an insoluble CSP, it always backjumps from a dead-end state; i.e., it does not terminate or backjump from a situation in which a solution of the CSP was found.

**Lemma 1** *Given an insoluble CSP and a variable ordering for CBJ, there exists a variable ordering for BT such that BT never visits more nodes than CBJ to show that no solution exists.*

**Proof** In the backtrack tree generated by CBJ under the variable ordering, let the last backjump that terminates the execution of CBJ be from variable $x_j$ to the root of the backtrack tree. We choose $x_j$ to be the first variable for BT. For each value $a$ in the domain of $x_j$, if the current node in the backtrack tree for CBJ is *consistent* (not a leaf node), the next variable chosen to be instantiated after assigning $a$ to $x_j$ is the variable that backjumps to $x_j$ and causes the assignment $x_j \leftarrow a$ to be revoked. The entire variable ordering for BT can be worked out in a similar, recursive manner. For this variable ordering for BT to be well-defined, it remains to show that if the current node in the backtrack tree for CBJ is inconsistent (a leaf node), the corresponding node in the backtrack tree for BT is also inconsistent (and therefore no next variable needs to be chosen). We show that the variables skipped in the variable ordering constructed for BT are irrelevant to the dead-end states encountered by CBJ. Suppose at a stage we have ordered the variables to be instantiated for BT as $x_{j_1}, \ldots, x_{j_k}$, and for value $a \in dom(x_{j_k})$ we choose the next variable $x_{j_{k+1}}$ as the variable which backjumps to the current variable $x_{j_k}$ in the CBJ backtrack tree. We prove by induction that the conflict set of $x_{j_{k+1}}$ used in the backjumping is subsumed by $\{x_{j_1}, \ldots, x_{j_k}\}$. $k = 1$ is the case of the last backjump that terminates the execution of CBJ. The hypothesis is true because the conflict set of $x_{j_1}$ is an empty set. Suppose it is true for the case of $k > 1$. Because $x_{j_{k+1}}$ backjumps to $x_{j_k}$, the conflict set of $x_{j_{k+1}}$ is merged in the conflict set of $x_{j_k}$. From the inductive assumption, the conflict set of $x_{j_k}$ is subsumed by $\{x_{j_1}, \ldots, x_{j_{k-1}}\}$, and thus the conflict set of $x_{j_{k+1}}$ is subsumed by $\{x_{j_1}, \ldots, x_{j_k}\}$. Therefore, the hypothesis holds for the case of $k+1$. If CBJ finds out that instantiation $x_{j_k} \leftarrow a$ is inconsistent with the assignments of some past variables which are added to the conflict set of $x_{j_k}$, BT is also able to find out the inconsistency because the conflict set of $x_{j_k}$ is subsumed by $\{x_{j_1}, \ldots, x_{j_{k-1}}\}$. Thus, the variable ordering for BT is well-defined. ∎

For soluble CSPs, we further distinguish the problem between finding one solution and finding all solutions.

**Lemma 2** *Given a CSP and a variable ordering for CBJ to find the first solution, there exists a variable ordering for BT such that BT never visits more nodes than CBJ to find the first solution.*

**Proof** Without loss of generality, let $\{x_1 \leftarrow a_1, \ldots, x_n \leftarrow a_n\}$ be the first solution found. A variable ordering for BT can be constructed in the following way. The first variable chosen for BT is $x_1$ as it is the first variable in the path from the root to the solution in the CBJ backtrack tree. Because we assume a static value ordering in the backtracking search, all values in the domain of $x_1$ that precede value $a_1$ must be rejected by CBJ and BT before value $a_1$ is used to instantiate $x_1$. Furthermore, because $\{x_1 \leftarrow a_1, \ldots, x_n \leftarrow a_n\}$ is the first solution encountered by CBJ under the above variable ordering and value ordering, the instantiation of $x_1$ with a value preceding $a_1$ leads to an insoluble subproblem and eventually CBJ backjumps from a deeper variable to $x_1$ to revoke that assignment. Note that $x_1$ cannot be skipped by a backjump from a deeper variable because $x_1$ is on the first level of the search tree and there is a solution for the CSP. Assigning $x_1$ with each of the values that precede $a_1$ in its domain leads to insoluble subproblems and the instantiation order for BT can be arranged as in Lemma 1. Whenever $x_k$ is instantiated with value $a_k$, $x_{k+1}$ is chosen to be the next variable, as it follows $x_k$ in the path from the root to the solution in the CBJ backtrack tree. Again, all values in the domain of $x_{k+1}$ that precede $a_{k+1}$ in the value ordering must be rejected by CBJ and BT before $a_{k+1}$ is assigned to $x_{k+1}$. The instantiation of $x_{k+1}$ with each of these values leads to an insoluble subproblem and eventually CBJ backjumps from a deeper variable to $x_{k+1}$. Similarly, $x_{k+1}$ cannot be skipped by a backjump from a deeper variable because otherwise at least one of the assignments to $x_1, \ldots, x_k$ must be changed so that $\{x_1 \leftarrow a_1, \ldots, x_n \leftarrow a_n\}$ is not the first solution encountered by CBJ. In each of these insoluble subproblems, the instantiation order for BT can be arranged as in Lemma 1. Finally, $x_n$ is instantiated with $a_n$ and BT finds the solution. ∎

When CBJ is used to find all solutions, special steps must be taken to handle the conflict sets. The problem here is that the conflict sets of CBJ are meant to indicate which instantiations are responsible for some previously discovered inconsistency. However, after a solution is found, conflict sets cannot always be interpreted in this way. It is the search for other solutions, rather than an inconsistency, that causes the algorithm to backtrack. We need to differentiate between two causes of CBJ backtracks: (1) detecting an inconsistency, and (2) searching for other solutions. In the latter case, the backtrack must be always chronological; that is, to the immediately preceding variable. A simple solution is to remember the number of solutions found so far when a variable is chosen to be instantiated, and later when a dead-end state is encountered at this level, we compare the recorded number with the current number of solutions. A difference indicates that some solutions have been found in this interval of search, and forces the algorithm to backtrack chronologically. Otherwise the algorithm performs a normal backjumping by analyzing the conflict set of the current variable.

**Lemma 3** *Given a CSP and a variable ordering for CBJ to find all solutions, there exists a variable ordering for BT such that BT never visits more nodes than CBJ to find all solutions.*

**Proof** Let the first solution found by CBJ be $\{x_1 \leftarrow a_1, \ldots, x_n \leftarrow a_n\}$ in the order of $x_1, \ldots, x_n$. We first construct the variable ordering for BT as it is applied to find the first solution. However, because BT follows a strict chronological backtracking, it will inevitably visit all the nodes $\{x_1 \leftarrow a_1, \ldots, x_{j-1} \leftarrow a_{j-1}, x_j \leftarrow a'_j\}$, where $1 \leq j \leq n$ and $a'_j$ comes after $a_j$ in the domain of $x_j$. If CBJ skips any of these nodes, for example, from a deeper level variable $x_h$ to $x_{j-1}$, while the instantiations of $x_1, \ldots, x_j$ have not been changed, BT will possibly visit more nodes than CBJ. We will show this cannot happen by induction on the distance between the current level $j$ and the deepest level $n$. After CBJ has found the solution at level $n$, it will try other values for $x_n$ and eventually backtrack to $x_{n-1}$. So the nodes at level $n$ cannot be skipped. Suppose it is true for the case of level $j+1$ and now we consider the case of level $j$. Because $x_j \leftarrow a_j$ was not skipped in the backjumping, if $a_j$ is the last value in its domain, CBJ will backtrack to $x_{j-1}$ because the number of solutions has been changed. So it is true for the case of $j$. Otherwise CBJ will change the instantiation of $x_j$ to the next value in its domain. Let the current partial solution be $t = \{x_1 \leftarrow a_1, \ldots, x_{j-1} \leftarrow a_{j-1}, x_j \leftarrow a'_j\}$. If the subtree rooted by $t$ contains solutions, from the inductive hypothesis, CBJ will not skip this node because it is on level $j$. If the subtree rooted by $t$ contains no solution, there exists a backjump from a deeper level variable $x_h$ to escape this subtree. Could it jump beyond $x_j$ such that $t$ is skipped? In that case, the conflict set of $x_h$ is subsumed in $\{x_1, \ldots, x_{j-1}\}$. From the definition of conflict set, we know that the current instantiations of the variables in the conflict set cannot lead to a solution. However the current instantiations of $\{x_1, \ldots, x_{j-1}\}$ do lead to a solution, $\{x_1 \leftarrow a_1, \ldots, x_n \leftarrow a_n\}$. That is a contradiction. So the conflict set of $x_h$ must contain $x_j$ and thus the node $t$ at level $j$ cannot be skipped. After all the values in the domain of $x_j$ have been tried, CBJ will chronologically backtrack to $x_{j-1}$ because the number of solutions has changed. Thus, $x_{j-1} \leftarrow a_{j-1}$ will not be skipped. The hypothesis is true for the case of any level $j$. Then we construct the variable ordering for BT in the following way: If the current partial solution $t = \{x_1 \leftarrow a_1, \ldots, x_{j-1} \leftarrow a_{j-1}, x_j \leftarrow a'_j\}$ cannot be extended to a solution, we construct a variable ordering for the insoluble subproblem. If $t$ can be extended to a solution, we construct a variable ordering for BT as the case of finding the first solution in this subproblem, and recursively apply the above steps until a backjump to level $x_j$ changes the instantiation $x_j \leftarrow a'_j$. Under the above variable ordering, BT will never visit more nodes than CBJ. ∎

**Theorem 4** *Given a CSP and a variable ordering for CBJ, there exists a variable ordering for BT such that BT never visits more nodes than CBJ in solving the CSP.*

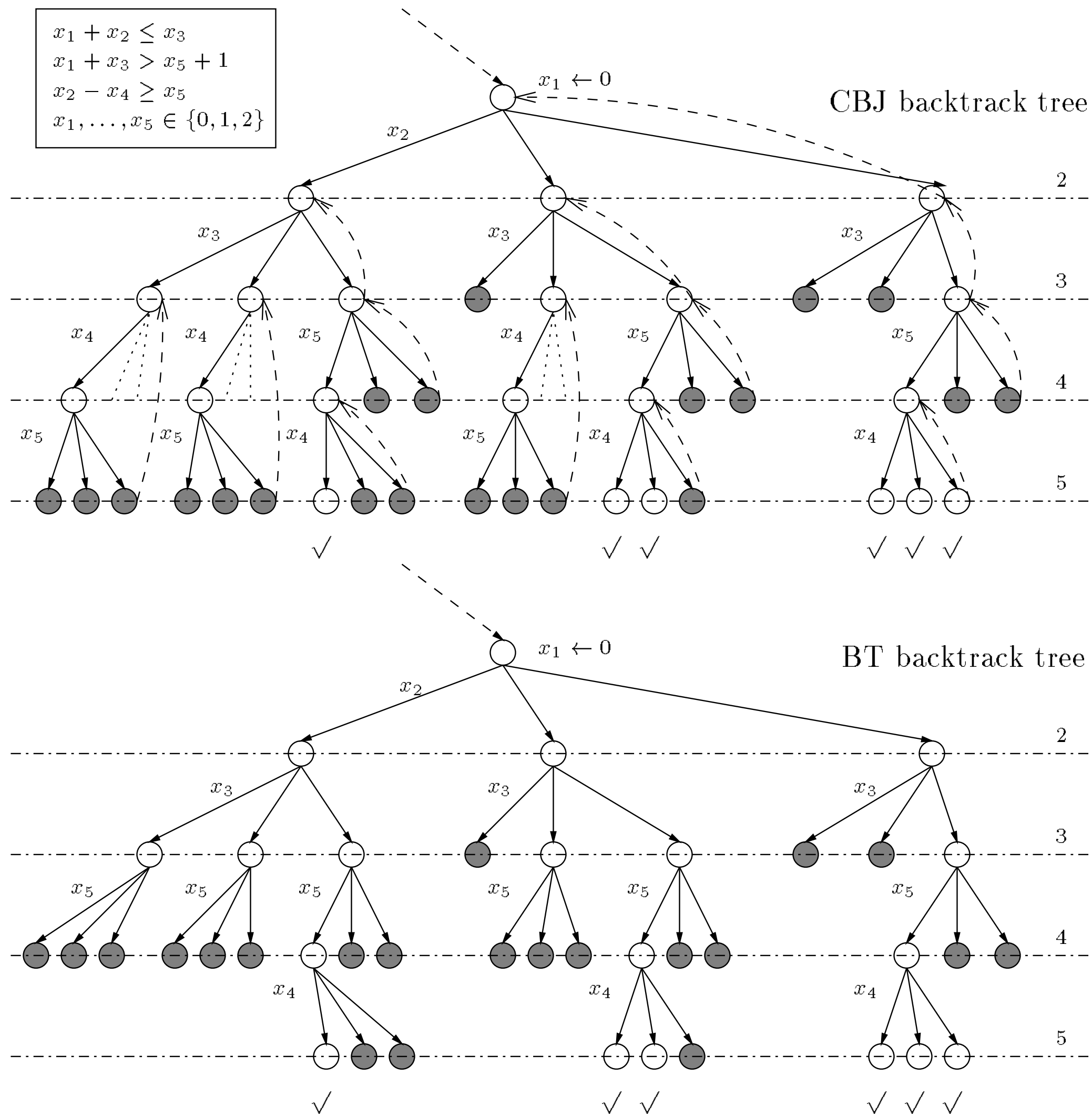

Figure 1: An illustration of the variable ordering constructed for BT from a CBJ backtrack tree (for the CSP shown upper left).

**Proof** Follows from Lemmas 1, 2, and 3. ∎

**Example 1** *Figure 1 shows the BT backtrack tree based on the variable ordering constructed from the execution of CBJ to solve a CSP under a (hypothetical) dynamic variable ordering. The first solution found by CBJ is* $\{x_1 \leftarrow 0, x_2 \leftarrow 0, x_3 \leftarrow 2, x_5 \leftarrow 0, x_4 \leftarrow 0\}$. *Thus, BT first instantiates* $x_1$ *and* $x_2$ *to 0. The node* $\{x_1 \leftarrow 0, x_2 \leftarrow 0, x_3 \leftarrow 0\}$ *and* $\{x_1 \leftarrow 0, x_2 \leftarrow 0, x_2 \leftarrow 1\}$ *in the CBJ backtrack tree lead to insoluble subproblems. The variable ordering for BT at each of these nodes is constructed as in the case of insoluble CSPs. For example, in the CBJ backtrack tree, the last backjump to revoke the node* $\{x_1 \leftarrow 0, x_2 \leftarrow 0, x_3 \leftarrow 0\}$ *is from* $x_5$ *to* $x_3$*, so the next variable instantiated in BT at this node is* $x_5$*. Under such an ordering, BT avoids instantiating* $x_4$ *and visits fewer nodes than CBJ. Then BT instantiates* $x_3$ *to 2,* $x_5$ *to 0, and* $x_4$ *to 0, and finds the first solution.*

We have shown that there exists a "perfect" variable ordering such that CBJ becomes redundant. Of course, the "perfect" ordering would not be known *a priori*, and in practice, the primary goal in designing variable ordering heuristics is not to simulate the execution of CBJ, but to reduce the size of the overall backtrack tree. As an example, the popular fail-first heuristic selects as the next variable to be instantiated the variable with the minimal remaining domain size (the size of the domain after removing values that are in conflict with past instantiations) as this can be shown to minimize the size of the overall tree under certain assumptions. A secondary effect, however, is that variables that have conflicts with past instantiations are likely to be instantiated sooner, thus approximating the "perfect" ordering and diminishing the effects of backjumping.

## 4. Maintaining Consistency and Backjumping

In this section, we present theoretical results that deepen our understanding of the relationship between the look-ahead technique of maintaining a level of local consistency during the backtracking search and the look-back technique of CBJ.

In previous work, Kondrak and van Beek (1997) show that, given the same deterministic static or dynamic variable ordering heuristic, CBJ never visits more nodes than BT and FC-CBJ never visits more nodes than FC. Prosser (1993a) shows that the removal of an inconsistent value from the domain of a variable can diminish the effects of CBJ and that CBJ can visit fewer nodes than an algorithm that combines CBJ with the discovery and removal of some inconsistent values. Previous empirical work shows that the number of nodes that CBJ saves depends on the level of local consistency maintained (Bacchus & van Run, 1995; Bessière & Régin, 1996; Prosser, 1993b).

We extend the partial ordering of backtracking algorithms presented by Kondrak and van Beek (1997) to include backtracking algorithms and their CBJ hybrids that maintain levels of local consistency beyond forward checking, including the important algorithms that maintain arc consistency. We show that CBJ and an algorithm that maintains strong $k$-consistency in the backtracking search are incomparable in that each can be exponentially better than the other. This result is refined by using the concept of *backjump level* in the execution of a backjumping algorithm and showing that an algorithm that maintains strong $k$-consistency never visits more nodes than a backjumping algorithm that is allowed to backjump at most $k$ levels. Thus, as the level of local consistency that is maintained in the backtracking search is increased, the less that backjumping will be an improvement.

In Section 4.1, we consider the backjumping algorithms and define the series of algorithms $\mathrm{BJ}_k$. In Section 4.2, we consider the look-ahead algorithms that maintain a level of local consistency and define the series of algorithms $\mathrm{MC}_k$. Finally, in Section 4.3, we consider the relationships between the backjumping and the look-ahead algorithms and their hybrids. The reader who is not interested in the technical proofs of the results should jump directly to this section.

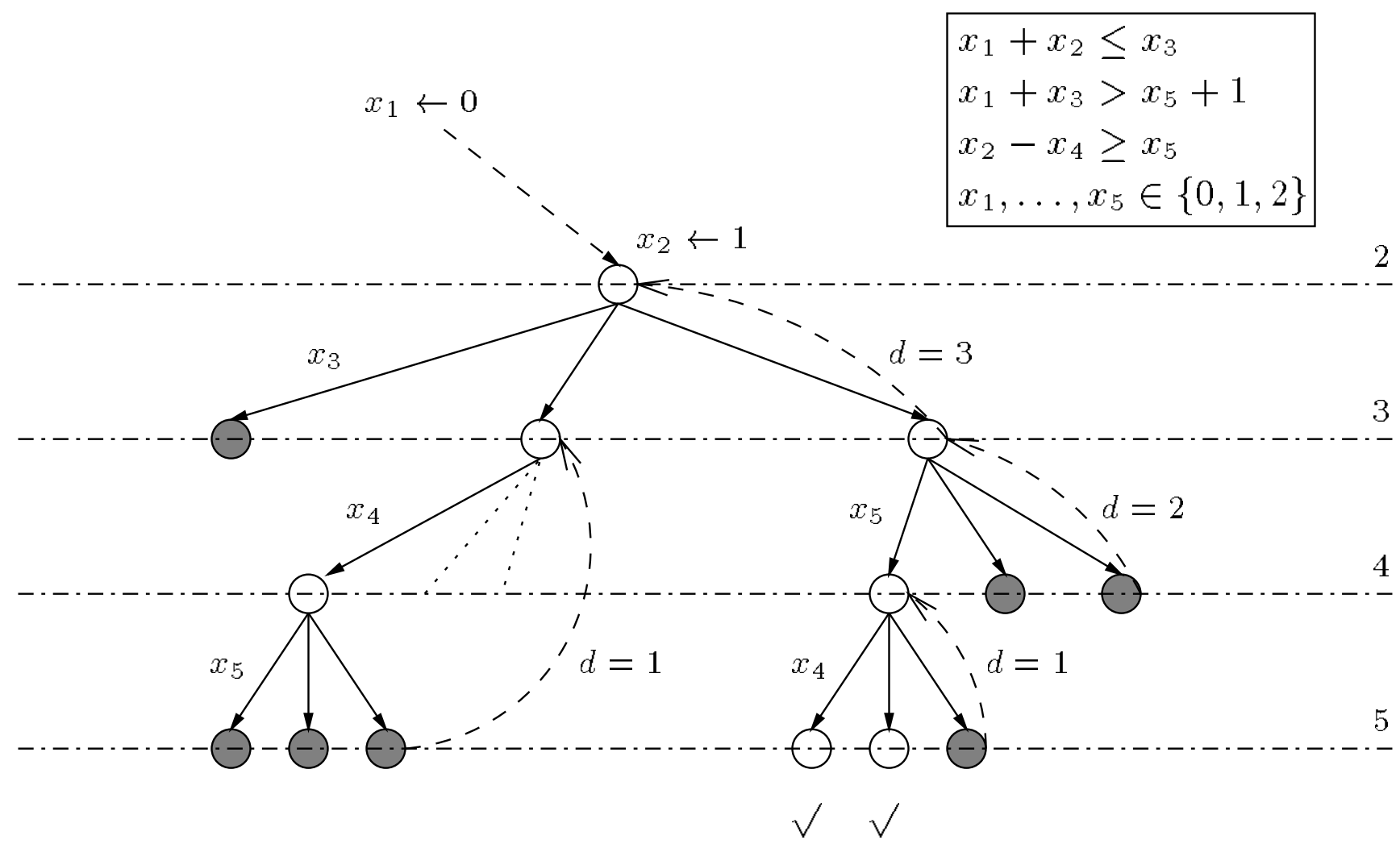

Figure 2: An illustration of backjump levels in a CBJ backtrack tree (for the CSP shown upper right).

### 4.1 Backjump Level and BJ$_k$

To analyze the influence of the level of consistency on the backjumping, we need the notion of *backjump level*. Informally, the level of a backjump is the distance, measured in backjumps, from the backjump destination to the "farthest" dead-end.

**Definition 6 (backjump level, Kondrak & van Beek, 1997)** *The definition of backjump level is recursive:*
*1. A backjump from variable $x_i$ to variable $x_h$ is of level 1 if it is performed directly from a dead-end state in which every value of $x_i$ fails a consistency check.*
*2. A backjump from variable $x_i$ to variable $x_h$ is of level $d \geq 2$, if all backjumps performed to variable $x_i$ are of level less than $d$, and at least one of them is of level $d - 1$.*

**Example 2** *Figure 2 shows the backjump levels in an example CBJ backtrack tree. There is a one-level backjump from $x_5$ to $x_3$ because every value in the domain of $x_5$ fails a consistency check. Then CBJ finds two solutions for the problem and thus it chronologically backtracks from $x_4$ to $x_5$, and later to $x_3$. The backjumps are of level one and two respectively. At last there is a three-level backjump from $x_3$ to $x_2$.*

By classifying the backjumps performed by a backjumping algorithm into different levels, we can now weaken CBJ into a series of backjumping algorithms which perform limited levels of backjumps. BJ$_k$ is a backjumping algorithm which is allowed to perform at most $k$-level backjumps and it chronologically backtracks when a $j$-level backjump for $j > k$ is encountered[3]. BJ$_n$ is equivalent to CBJ, which performs unlimited backjumps, and BJ$_1$ is

3. BJ$_k$ is only of theoretical interest since in practice one would use CBJ rather than artificially prevent backjumping; i.e., one has to actually add code to prevent backjumping.

equivalent to Gaschnig's (1978) BJ, which only does the first level backjumps or backjumps from dead-ends.

One may immediately conclude that $\text{BJ}_{k+1}$ is always better than $\text{BJ}_k$ because it does one more level of backjumps. However, to be more precise, we need to justify that a situation where $\text{BJ}_k$ may skip a node visited by $\text{BJ}_{k+1}$ does not exist. Similar to a result by Kondrak and van Beek (Theorem 11, 1997), we can show that:

**Theorem 5** *$BJ_k$ visits all the nodes that $BJ_{k+1}$ visits.*

### 4.2 Maintaining Strong $k$-consistency ($\text{MC}_k$)

Although backtracking algorithms that maintain arc consistency (or a truncated form of arc consistency called forward checking) during the search have been well-studied, a backtracking algorithm that maintains strong $k$-consistency ($\text{MC}_k$) has never been fully addressed in the literature. In order to study the relationship between $\text{BJ}_k$ and $\text{MC}_k$, we need to specify precisely the $\text{MC}_k$ algorithms.

A generic scheme to maintain a level of local consistency in a backtracking search is to perform at each node in the search tree one full cycle of consistency achievement. A consistency achievement algorithm is applied to the CSP which is *induced* by the current partial solution. If, as a result, the induced CSP becomes empty after applying the consistency algorithm, the instantiation of the current variable is a dead-end and should be rejected. If the resulting CSP is not empty, the instantiation of the current variable is accepted and the search continues to the next level.

The simplest form of an induced CSP is to restrict the domains of the instantiated variables to have only one value and leave the set of constraints unchanged. This idea can be traced back to Gaschnig's (1978) implementation of MAC, referred to as DEEB; i.e., Domain Element Elimination with Backtracking. However, in order to establish a relation between $\text{BJ}_k$ and $\text{MC}_k$, we need a more restricted definition of the induced CSP, where the constraints in the induced CSP are the selections and projections of the constraints in the original CSP with respect to a partial solution.

**Definition 7 (induced CSP)** *Given a consistent partial solution $t$ of a CSP $P$, the CSP induced by $t$, denoted by $P|_t$, has all the variables in $P$ except those instantiated by $t$, the domain of each variable is the same as in $P$, and for each constraint $C$ in $P$ where $vars(C) \nsubseteq vars(t)$, there is a constraint $C' = \pi_{vars(C)-vars(t)}(\sigma_{t[vars(C)\cap vars(t)]}(C))$ in $P|_t$.*

**Example 3** *Consider the graph coloring problem and the corresponding CSP shown in Figure 3. The original CSP has four variables, $x_1, \ldots, x_4$, where $x_1, x_2, x_3 \in \{r, g, b\}$ and $x_4 \in \{r\}$, and five binary constraints, $x_1 \neq x_2$, $x_1 \neq x_3$, $x_2 \neq x_3$, $x_2 \neq x_4$ and $x_3 \neq x_4$. Given a partial solution $t = \{x_1 \leftarrow g, x_2 \leftarrow b\}$, the CSP induced by $t$, $P|_t$, has two variables, $x_3$ and $x_4$, and the unary and binary constraints shown in Figure 4.*

The maintaining strong $k$-consistency algorithm ($\text{MC}_k$) at each node in the backtrack tree applies a strong $k$-consistency achievement algorithm to the CSP induced by the current partial solution. Under such an architecture, FC can be viewed as maintaining one-consistency, and for binary CSPs, MAC can be viewed as maintaining strong two-consistency.

An algorithm enforcing strong $k$-consistency on a CSP instance should detect and remove all those inconsistencies $t = \{x_1 \leftarrow a_1, \ldots, x_j \leftarrow a_{j-1}\}$ where $1 \leq j \leq k$ and $t$ is consistent but cannot be consistently extended to some $j^{th}$ variable $x_j$. To remove an inconsistency, we make it inconsistent in the resulting CSP by removing values from domains, removing inconsistent tuples from existing constraints, or adding new constraints to the CSP.

We use the concept of a $k$-proof-tree in characterizing the tuples that are removed by a strong $k$-consistency achievement algorithm.

**Definition 8 ($k$-proof-tree)** *A $k$-proof-tree for a partial solution $t$ over at most $k$ variables in a CSP is a tree in which each node is associated with a partial solution over at most $k$ variables in the CSP, where (1) the root of the $k$-proof-tree is associated with $t$, and (2) each leaf node of the $k$-proof-tree is inconsistent in the CSP, and (3) each non-leaf node $s$ of the $k$-proof-tree is consistent in the CSP, and the children of $s$ at the next level are nodes $s' \cup \{x \leftarrow a_1\}, \ldots, s' \cup \{x \leftarrow a_l\}$ such that $s' \subseteq s$, $x \notin vars(s)$, and $dom(x) = \{a_1, \ldots, a_l\}$.*

**Example 4** *Figure 3 shows a three-proof-tree (more than one is possible) for $t = \{x_1 \leftarrow g\}$ in the given graph coloring problem. Each non-leaf node, including the root $t$, is consistent, and each leaf node is inconsistent in the CSP. Since we have constructed a three-proof-tree for the tuple $t$ it cannot be part of a solution to the CSP and a strong 3-consistency achievement algorithm would remove it.*

In general, if a $k$-proof-tree for an inconsistency in a CSP can be constructed, an algorithm achieving strong $k$-consistency would deduce and remove the inconsistency. After applying a strong $k$-consistency achievement algorithm on the CSP, if all the children of a node in the $k$-proof-tree are inconsistent in the resulting CSP, that node is also inconsistent in the resulting CSP because one of its subtuples cannot be consistently extended to an additional variable. Because all the leaf nodes in the $k$-proof-tree are inconsistent in the original CSP, in a bottom-up manner the inconsistency of the root of the tree can be deduced and removed from the resulting CSP. As a special case, if a $k$-proof-tree for the empty inconsistency in a CSP can be constructed, the CSP will be empty after enforcing strong $k$-consistency since every way to extend a variable has been shown to lead to an inconsistency (and therefore, each value would be removed from the domain resulting in the empty domain). On the other hand, after a CSP has been made strongly $k$-consistent, if a partial solution $t$ over at most $k$ variables is inconsistent in the resulting CSP, a $k$-proof-tree for $t$ in the original CSP can be constructed. If $t$ is inconsistent in the original CSP, the $k$-proof-tree contains the single node $t$. Otherwise, $t$ or a subtuple $t'$ of $t$ cannot be extended to an additional variable $x$; i.e., all the partial solutions $t' \cup \{x \leftarrow a_1\}, \ldots, t' \cup \{x \leftarrow a_l\}$, where $dom(x) = \{a_1, \ldots, a_l\}$, are inconsistent in the resulting CSP. Then we can construct the $k$-proof-tree recursively for each of those inconsistencies. As a special case, if a CSP is empty after enforcing strong $k$-consistency, a $k$-proof-tree for the empty inconsistency in the original CSP can be constructed.

The following lemmas (Lemma 6 to Lemma 8) reveal some basic properties about induced CSPs and strong $k$-consistency enforcement on induced CSPs, which are used in the proofs of Theorem 10 and Theorem 14.

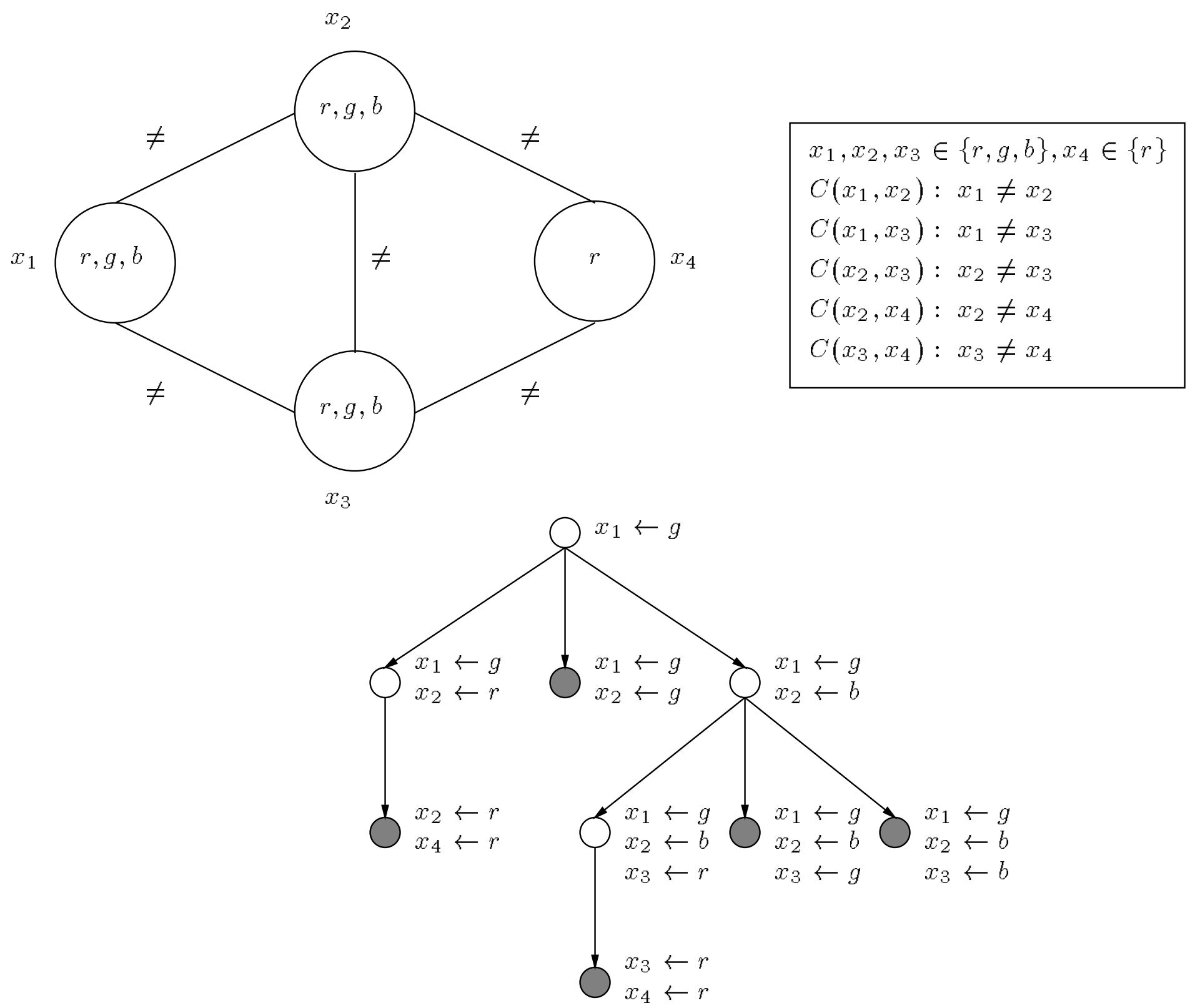

Figure 3: A three-proof-tree for $\{x_1 \leftarrow g\}$ in the graph coloring problem. All leaf nodes in the proof-tree are inconsistent in the CSP.

**Lemma 6** *Given a CSP $P$ and two partial solutions $t$ and $t'$ of $P$, if $t \subset t'$, then $P|_{t'} = (P|_t)|_{t'-t}$.*

**Proof** Clearly $P|_{t'}$ and $(P|_t)|_{t'-t}$ have the same set of variables and the same set of domains. Because $\pi_{vars(C)-vars(t')}\sigma_{t'}C = \pi_{vars(C)-vars(t')}\sigma_{t'-t}(\pi_{vars(C)-vars(t)}\sigma_t C)$, for each constraint $C$ in $P$, the same selection and projection are made in $P|_{t'}$ and $(P|_t)|_{t'-t}$. Therefore, $P|_{t'}$ and $(P|_t)|_{t'-t}$ have the same set of constraints. ∎

**Lemma 7** *Given a CSP $P$ and a consistent partial solution $t$ of $P$, if (i) $P$ is empty after achieving strong $k$-consistency, or (ii) there exists a variable $x \in vars(t)$ such that the value $t[x]$ is removed from the domain of $x$ when achieving strong $k$-consistency on $P$, then $P|_t$ is empty after achieving strong $k$-consistency,*

**Proof** We first show that, given a consistent partial solution $t$ of a CSP $P$, and a $k$-proof-tree $T$ for an inconsistency $s$ in $P$, there is a corresponding well-defined $k$-proof-tree $T_t$ for the inconsistency $s' = s[vars(s) - vars(t)]$, in the induced CSP $P|_t$, provided $s$ does not contain any assignments that are inconsistent with the assignments in $t$. $T_t$ is constructed from $T$ in three steps (see Figure 4 for an example): (Step 1) Remove from $T$ all nodes and their descendants which contain assignments that are inconsistent with the assignments in $t$. (Step 2) Replace each remaining node $t'$ in $T$ with the node $t'' = t'[vars(t') - vars(t)]$; i.e., remove those variables which occur in $t$ and thus do not occur in $P|_t$. If $t'$ is not a leaf node in $T$, then by definition $t'$ is consistent in $P$. It is possible that the corresponding node $t''$ in $T_t$ is inconsistent in $P|_t$. Should this be the case, we make $t''$ into a leaf node by removing all of its descendants. If $t'$ is a leaf node in $T$, then by definition $t'$ is inconsistent in $P$; i.e., there exists a constraint $C$ in $P$ such that $t'$ does not satisfy $C$. It must be the case that $vars(C) \not\subseteq vars(t)$ (since $vars(C) \subseteq vars(t)$ contradicts the fact that $t'$ is inconsistent with $C$ and $t$ is consistent and therefore consistent with $C$, but $t'$ and $t$ agree on their assignments by Step 1). Hence, there is a corresponding constraint $C'$ in $P|_t$ which is the selection and projection of $C$ in $P$. Now, it is easy to verify that the corresponding node $t''$ is also inconsistent with $C'$ and is therefore a well-defined leaf node. (Step 3) Remove all subsumed nodes from $T$, where node $t_2$ is subsumed by node $t_1$ if $t_2$ is a (necessarily only) child of $t_1$ and $vars(t_2) \subseteq vars(t_1)$. All children of a subsumed node $t_2$ are made children of the parent of $t_2$.

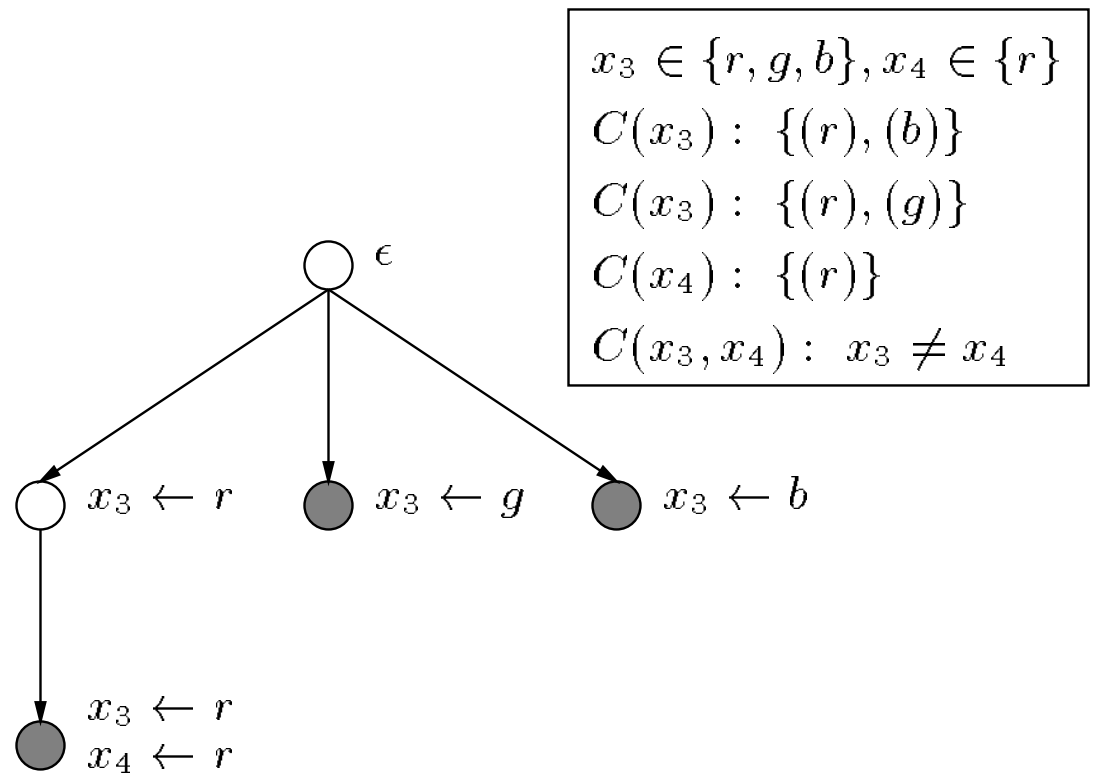

Figure 4: Proof-tree for the empty inconsistency in the CSP $P|_t$ induced by $t = \{x_1 \leftarrow g, x_2 \leftarrow b\}$ constructed from the proof-tree for $\{x_1 \leftarrow g\}$ in the CSP $P$ shown in Figure 3.

Now, suppose $P$ is empty after achieving strong $k$-consistency. Then there is a $k$-proof-tree for the empty inconsistency in $P$ and we can construct a $k$-proof-tree for the empty inconsistency in $P|_t$. Therefore, $P|_t$ is empty after achieving strong $k$-consistency. Suppose there exists a variable $x \in vars(t)$, such that the value $t[x]$ is removed from the domain of $x$ when achieving strong $k$-consistency on $P$. Then there is a $k$-proof-tree for $\{x \leftarrow t[x]\}$ in $P$ and we can construct a $k$-proof-tree for the empty inconsistency in $P|_t$. Therefore $P|_t$ is empty after achieving strong $k$-consistency. ∎

**Lemma 8** *Given a CSP P and an assignment* $\{x \leftarrow a\}$, $a \in dom(x)$, *if the induced CSP* $P|_{\{x \leftarrow a\}}$ *is empty after achieving strong* $(k-1)$*-consistency, then the value a is removed from the domain of x when achieving strong k-consistency on P.*

**Proof** Suppose $P|_{\{x \leftarrow a\}}$ is empty after achieving strong $(k-1)$-consistency. Thus, there is a $(k-1)$-proof-tree for the empty inconsistency in $P|_{\{x \leftarrow a\}}$. We now convert the $(k-1)$-proof-tree to a $k$-proof-tree for $\{x \leftarrow a\}$ in $P$. Each node $t$ in the original $(k-1)$-proof-tree is replaced by $t \cup \{x \leftarrow a\}$. Thus, the root of the tree becomes $\{x \leftarrow a\}$. Furthermore, if $t$ is not a leaf node in the original $(k-1)$-proof-tree; i.e., $t$ is consistent in $P|_{\{x \leftarrow a\}}$, it is easy to verify that $t \cup \{x \leftarrow a\}$ is consistent in $P$. If $t$ is a leaf node in the original $(k-1)$-proof-tree; i.e., $t$ is inconsistent in $P|_{\{x \leftarrow a\}}$, there is a constraint $C'$ in $P|_{\{x \leftarrow a\}}$ such that $t$ does not satisfy $C'$. Let $C'$ be the selection and projection of the constraint $C$ in $P$. Thus, $t \cup \{x \leftarrow a\}$ does not satisfy the constraint $C$ in $P$ and is therefore inconsistent in $P$. Hence, we have constructed a $k$-proof-tree for $\{x \leftarrow a\}$ in $P$ and thus $a$ would be removed from the domain of $x$ when achieving strong $k$-consistency on $P$. ∎

$MC_k$ extends the current node if the CSP induced by the current partial solution is not empty after achieving strong $k$-consistency. The node is thus called a *k-consistent node.*

**Definition 9 ($k$-consistent node)** *A node t in the search tree is a k*-consistent node *if the CSP induced by t is not empty after enforcing strong k-consistency. A node which is not k-consistent is called k-inconsistent.*

**Lemma 9** *If node t is k-consistent, its ancestors are also k-consistent.*

**Proof** Let $t'$ be one of $t$'s ancestors. Because $t' \subset t$, from Lemma 6, $P|_t = (P|_{t'})|_{t-t'}$. Thus, $P|_t$ is an induced subproblem of $P|_{t'}$. From Lemma 7, if $P|_t$ is not empty after achieving strong $k$-consistency, $P|_{t'}$ is not empty either after achieving strong $k$-consistency. Thus, $t'$ is $k$-consistent. ∎

The following theorem applies to the case of finding all solutions.

**Theorem 10** *If* $MC_k$ *visits a node, then its parent is k-consistent. If a node is k-consistent, then* $MC_k$ *visits the node.*

**Proof** The first part is true because $MC_k$ would not branch on this node if its parent was found $k$-inconsistent. We prove the second part by induction on the depth of the search tree. The hypothesis is trivial for $j = 1$. Suppose it is true for $j > 1$ and we have a $k$-consistent node $t$ at level $j+1$. Let the current variable be $x$. From Lemma 9, $t$'s parent $t'$ at level $j$ is $k$-consistent. Thus, $MC_k$ will visit $t'$. From Lemma 6, $P|_t = (P|_{t'})|_{\{x \leftarrow t[x]\}}$. Because $(P|_{t'})|_{\{x \leftarrow t[x]\}}$ is not empty after achieving strong $k$-consistency, from Lemma 7, value $t[x]$ will not be removed from the domain of $x$ when achieving strong $k$-consistency in $P|_{t'}$. As a consequence, $MC_k$ will visit $t$. ∎

A necessary and sufficient condition for $MC_k$ to visit a node $t$ is that $t$'s parent is $k$-consistent and the value assigned to the current variable by $t$ has not been removed from its domain when enforcing strong $k$-consistency on $t$'s parent.

**Theorem 11** *Given a CSP and a variable ordering, $MC_k$ visits all the nodes that $MC_{k+1}$ visits.*

**Proof** Follows from Theorem 10 and Lemma 7. ∎

### 4.3 Relationship Between $BJ_k$ and $MC_k$

Kondrak and van Beek (1997) have shown that for binary CSPs, BJ ($BJ_1$) visits all the nodes that FC ($MC_1$) visits, and FC-CBJ ($MC_1$-CBJ) and CBJ are incomparable. We extend their partial ordering of backtracking algorithms to include the relationship between $MC_k$, $BJ_k$, and $MC_k$-CBJ, $1 \leq k \leq n$. All of our results are for the case of general CSPs; i.e., they are not restricted to binary CSPs.

We begin by characterizing an important property of the CBJ algorithm.

**Lemma 12** *If CBJ performs a one-level backjump from a deeper variable $x_i$ to a shallower variable $x_h$, the node $t_h$ at the level of $x_h$ is one-inconsistent.*

**Proof** Let $S_i$ be the conflict set of $x_i$ used in the backjumping in which $x_h$ is the deepest variable. We show that $x_i$ will experience a domain wipe out when enforcing one-consistency on the induced CSP $P|_{t_h[S_i]}$. Each node $t_i$ at the level of $x_i$ is a leaf node; i.e., $t_i$ is inconsistent in $P$. Suppose $t_i$ does not satisfy constraint $C$ where $x_i \in vars(C)$ and $vars(C) \subseteq S_i \cup \{x_i\}$. The selection of $C$ in $P|_{t_h[S_i]}$, which constrains only one variable $\{x_i\}$, should prohibit value $t_i[x_i]$ of $x_i$. Thus, $x_i$ will experience a domain wipe out when enforcing one-consistency on $P|_{t_h[S_i]}$. Note that $P|_{t_h}$ is an induced subproblem of $P|_{t_h[S_i]}$. From Lemma 7, $P|_{t_h}$ is empty after enforcing one-consistency. Thus, $t_h$ at the level of $x_h$ is one-inconsistent. ∎

**Lemma 13** *If CBJ performs a $k$-level backjump from a deeper variable $x_i$ to a shallower variable $x_h$, the current node $t_h$ at the level of $x_h$ is $k$-inconsistent.*

**Proof** Let $S_i$ be the current conflict set of $x_i$ in which $x_h$ is the deepest variable. We show that if there is a $k$-level backjump from $x_i$ to $x_h$, then $P|_{t_h[S_i]}$ is empty after enforcing strong $k$-consistency and thus $t_h$ is $k$-inconsistent. The proof is by induction on $k$. $k = 1$ is true from Lemma 12. Suppose the hypothesis is true for the case of $k-1$ but it is not true for the case of $k$; i.e., there is a $k$-level backjump from $x_i$ to $x_h$, but the induced CSP $P|_{t_h[S_i]}$ is not empty after enforcing strong $k$-consistency. So there is at least one value $a$ left in the domain of $x_i$ after enforcing strong $k$-consistency on $P|_{t_h[S_i]}$. We know that the node $t_i$ at the level of $x_i$ instantiating $x_i$ with $a$ is either incompatible with $t_h$ (i.e., it is a leaf node) or is $l$-level backjumped from some deeper variable $x_j$, for some $1 \leq l < k$ (see Figure 5). However, $t_i$ cannot be a leaf node as otherwise $a$ would be removed from the domain of $x_i$ when enforcing strong $k$-consistency. Let $S_j$ be the conflict set of $x_j$. From the hypothesis, the induced CSP $P|_{t_i[S_j]}$ is empty after achieving strong $l$-consistency. Because value $a$ is not removed from the resulting CSP, from Lemma 8, the induced CSP $P|_{t_h[S_i] \cup \{x_i \leftarrow a\}}$ is not empty after achieving strong $(k-1)$-consistency. Because $t_i[S_j] \subseteq t_h[S_i] \cup \{x_i \leftarrow a\}$, the induced CSP $P|_{t_i[S_j]}$ is not empty after achieving strong $(k-1)$-consistency. That leads to a contradiction. Thus $P|_{t_h[S_i]}$ is empty after achieving strong $k$-consistency and $t_h$ at the level of $x_h$ is $k$-inconsistent. ∎

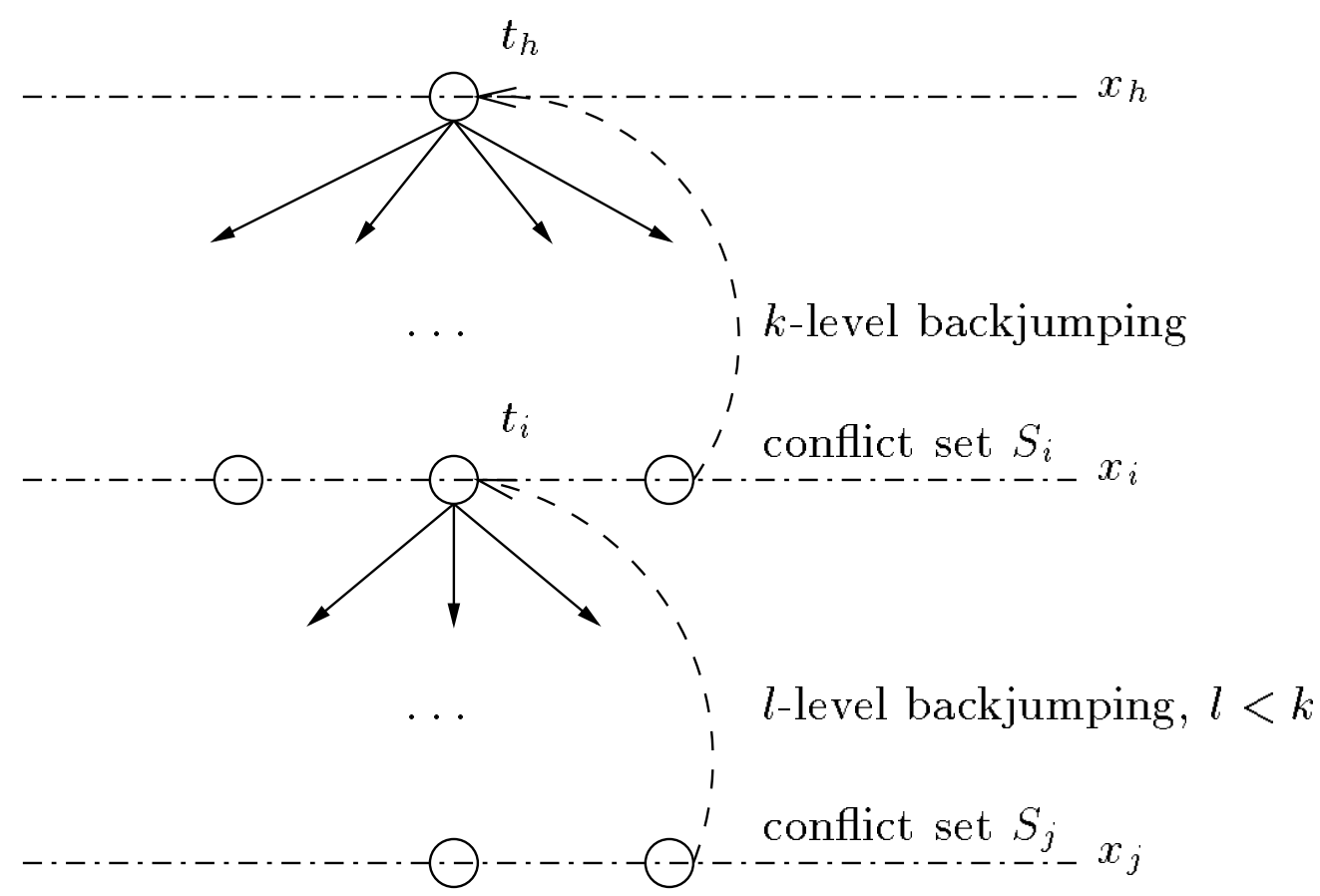

Figure 5: A scenario in the CBJ backtrack tree used in the proof of Lemma 13.

**Theorem 14** *Given a CSP and a variable ordering, $BJ_k$ visits all the nodes that $MC_k$ visits.*

**Proof** The proof is by induction on the level of the search tree. If $\mathrm{MC}_k$ visits a node at level $j$ in the search tree, $\mathrm{BJ}_k$ visits the same node. $j = 1$ is trivial. Suppose that it is true for the case of $j > 1$ and we have a node $t$ visited by $\mathrm{MC}_k$ at level $j + 1$. We know both $\mathrm{MC}_k$ and $\mathrm{BJ}_k$ visit $t$'s parent at level $j$. The only chance that $t$ may be skipped by $\mathrm{BJ}_k$ is that $\mathrm{BJ}_k$ backjumps from some deeper variable $x_i$ at level $i$ to a shallower variable $x_h$ at level $h$, such that $h < j+1 < i$. Thus, the node at level $h$ is $k$-inconsistent (by Lemma 13). Since the node at level $h$ is an ancestor of $t$ and we know $t$'s parent is $k$-consistent from Lemma 9, the node at level $h$ is $k$-consistent. That is the contradiction. Therefore, $\mathrm{BJ}_k$ visits $t$ at level $j + 1$. ∎

$\mathrm{MC}_k$ can be combined with backjumping, namely $\mathrm{MC}_k$-CBJ, provided the conflict sets are computed correctly after achieving strong $k$-consistency on the induced CSPs.

**Theorem 15** *Given a CSP and a variable ordering, $MC_k$ visits all the nodes that $MC_k$-CBJ visits.*

**Proof** Because $\mathrm{MC}_k$-CBJ behaves exactly the same as $\mathrm{MC}_k$ in the forward phase of a backtracking search, it is easy to verify that $\mathrm{MC}_k$-CBJ visits a node $t$ only if $t$'s parent is $k$-consistent and the value assigned to the current variable by $t$ was not removed from its domain when achieving strong $k$-consistency on $t$'s parent. Therefore, $\mathrm{MC}_k$-CBJ never visits more nodes than $\mathrm{MC}_k$ does. ∎

In Figure 6, we present a hierarchy in terms of the size of the backtrack tree for $\mathrm{BJ}_k$, $\mathrm{MC}_k$, and $\mathrm{MC}_k$-CBJ. If there is a path from algorithm $\mathcal{A}$ to algorithm $\mathcal{B}$ in the figure, we know that $\mathcal{A}$ never visits more nodes than $\mathcal{B}$ does. For example, $\mathrm{MC}_k$ never visits more nodes than $\mathrm{BJ}_j$, for all $j \leq k$. Otherwise, there are instances to show $\mathcal{A}$ may be exponentially better than $\mathcal{B}$, and *vice versa*.

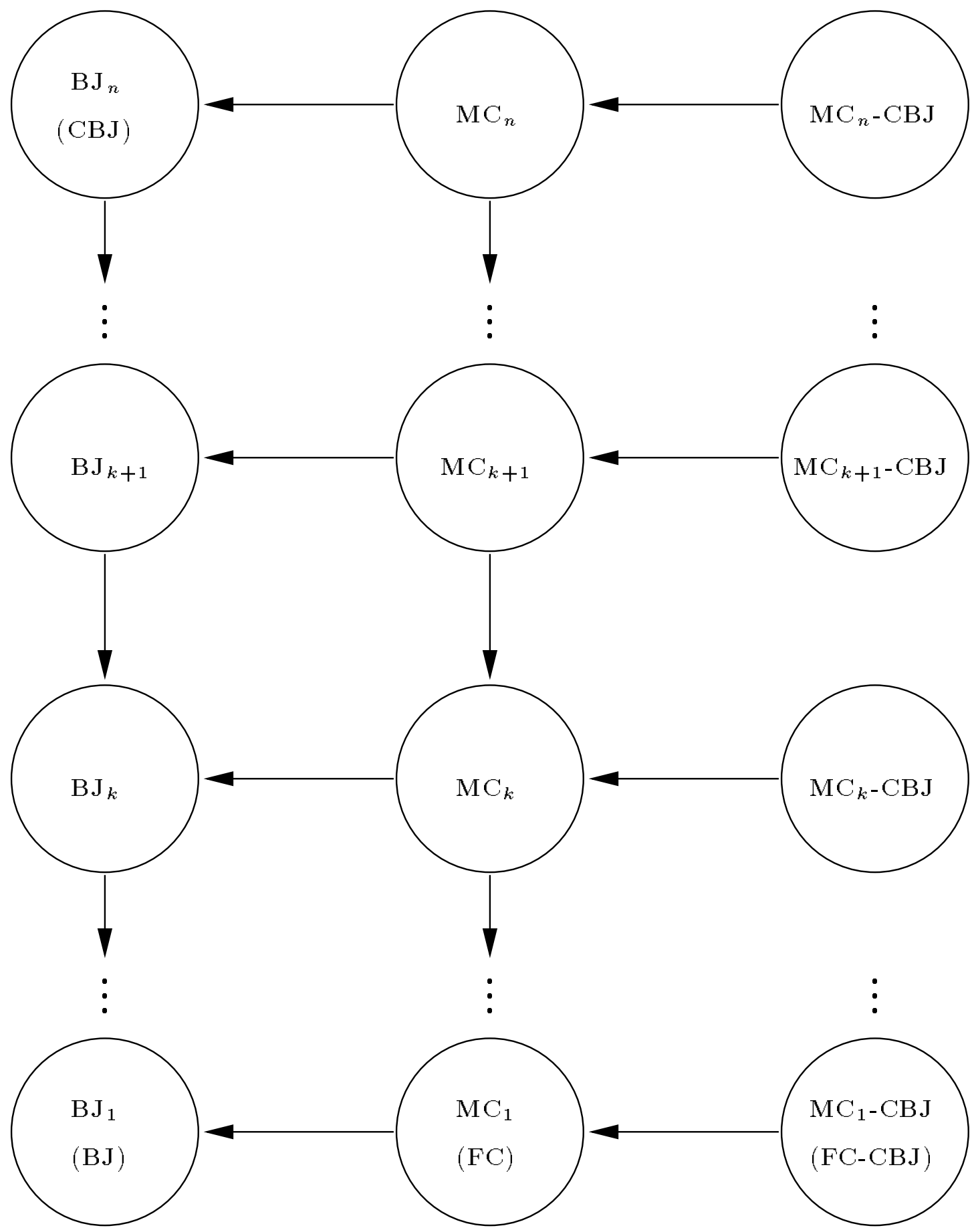

Figure 6: A hierarchy for $BJ_k$, $MC_k$, and $MC_k$-CBJ in terms of the size of the backtrack tree.

As the following example shows, for any fixed integer $k < n$, there exists a CSP instance such that CBJ visits exponentially fewer nodes than an algorithm that maintains strong $k$-consistency in the backtracking search[4].

**Example 5** *Given a fixed integer $k$, we can construct a binary CSP with $n+k+2$ variables, $x_1, \ldots, x_{n-k+1}, y_1, \ldots, y_{k+1}, x_{n-k+2}, \ldots, x_{n+1}$, where $dom(x_i) = \{1, \ldots, n\}$ for $1 \leq i \leq n+1$ and $dom(y_j) = \{1, \ldots, k\}$ for $1 \leq j \leq k+1$. The constraints are: $x_i \neq x_j$, for $i \neq j$, and $y_i \neq y_j$, for $i \neq j$. The problem consists of two separate pigeon-hole subproblems, one over variables $x_1, \ldots, x_{n+1}$ and the other over variables $y_1, \ldots, y_{k+1}$, and is insoluble. As we can see, the pigeon-hole problem is highly locally consistent. The first subproblem is strongly $n$-consistent and the second is strongly $k$-consistent. Under the above static variable ordering,*

4. Independently, Bacchus and Grove (1999) present a similar example to show that given a fixed $k$, CBJ may be exponentially better than an algorithm called $MIkC$, which essentially maintains $k$-consistency in the backtracking search.

*a backtracking algorithm maintaining strong $k$-consistency would not encounter a dead-end until $x_{n-k+1}$ is instantiated. Then it would find that the subproblem of $x_{n-k+2}, \ldots, x_{n+1}$ is not strongly $k$-consistent. Thus, the algorithm will backtrack before it reaches the second pigeon-hole subproblem. It will explore $\frac{n!}{k!}$ nodes at level $n-k+1$ of the search tree and thus take an exponential number of steps to find the problem is insoluble. CBJ does not encounter a dead-end at the level of $x_{n-k+1}$ and it continues to the second pigeon-hole problem. Eventually it will find the second-pigeon hole problem is insoluble and backjump to the root of the search tree. The total number of nodes explored is bounded by a constant, $O((k+1)^k)$, for a fixed $k$. Therefore, CBJ can be exponentially better than an algorithm maintaining strong $k$-consistency.*

Example 5 also shows that, although $\mathrm{MC}_k$ visits fewer nodes than $\mathrm{BJ}_k$ by Theorem 14, $\mathrm{BJ}_{k+1}$ can be exponentially better than $\mathrm{MC}_k$. However, $\mathrm{BJ}_{k+1}$ can be better than $\mathrm{MC}_k$ only if there is a $(k+1)$-level backjump that is not also a chronological backtrack. To see that this is true, suppose that on a particular instance all $(k+1)$-level backjumps are also chronological backtracks (i.e., the backjump is to the immediately preceding variable in the variable ordering and only that single variable becomes uninstantiated and is removed from the current partial solution). In this case, the freedom to backjump one additional level rather than chronologically backtrack does not make a difference and $\mathrm{BJ}_{k+1}$ is effectively $\mathrm{BJ}_k$ and thus cannot be better than $\mathrm{MC}_k$. Thus, $\mathrm{BJ}_{k+1}$ can be better than $\mathrm{MC}_k$ only if there is a $(k+1)$-level non-chronological backjump. We note, however, that since the number of backjumps of level $k+1$ is less than or equal to the number of backjumps of level $k$, as $k$ increases this gets more and more unlikely. Thus, as the level of local consistency that is maintained in the backtracking search is increased, the less that backjumping will be an improvement.

Consider Example 5 again. At each level of the backtrack tree for $\mathrm{MC}_k$, the instantiation of each of the past variables removes one distinct value from the domain of the current variable (recall that $\mathrm{MC}_k$ never instantiates the variable $y_1$ as it reaches a dead-end at $x_{n-k+1}$). If we were to maintain conflict sets for the variables, the conflict set for the current variable would include all of its past variables and thus when a dead-end is encountered by the algorithm, any backjump computed from the conflict sets would also necessarily be a chronologically backtrack. Thus, as this example shows, $\mathrm{MC}_k$-CBJ and $\mathrm{MC}_k$ can visit exactly the same nodes and consequently $\mathrm{BJ}_{k+1}$ can be exponentially better than $\mathrm{MC}_k$-CBJ. Furthermore, because $\mathrm{MC}_{k-1}$-CBJ can reach the second pigeon-hole problem without encountering a dead-end, it can finally retreat from the second pigeon-hole problem to the root of the search tree by backjumps. Thus, $\mathrm{MC}_{k-1}$-CBJ may be exponentially better than $\mathrm{MC}_k$-CBJ. In particular, this shows the surprising result that MAC-CBJ can visit exponentially more nodes than FC-CBJ.

Finally, as the following example shows, for any fixed integer $k < n$, there exists a CSP instance such that an algorithm that maintains strong $k$-consistency in the backtracking search visits exponentially fewer nodes than CBJ.

**Example 6** *Consider the CSP as defined in Example 5, but searched with the static variable ordering $y_1, \ldots, y_k, x_1, \ldots, x_{n+1}, y_{k+1}$.*

## 5. Empirical Evaluation of Adding CBJ to GAC

In this section, we report on experiments that examined the effect of adding CBJ to a backtracking algorithm that maintains generalized arc consistency (GAC), an algorithm that we refer to as GAC-CBJ. Previous work has shown the importance of algorithms that maintain arc consistency (e.g., Sabin & Freuder, 1994; Bessière & Régin, 1996). We show that adding CBJ to a backtracking algorithm that maintains generalized arc consistency can speed up the algorithm by several orders of magnitude on hard, structured problems.

Previous empirical studies of adding CBJ to a backtracking algorithm that maintains a level of local consistency have led to mixed conclusions. Adding CBJ to forward checking, a truncated form of arc consistency, has been shown to give improvements but not always significant ones. Prosser (1993b) observes that with a static variable ordering, FC-CBJ is about three times faster than FC on the Zebra problem. Smith and Grant (1995) observe that with a dynamic variable ordering, adding CBJ to FC led to significant savings but only on hard random problems that occur in the easy region. Bacchus and van Run (1995) observe that with a dynamic variable ordering, adding CBJ to FC only led to at most a 5% improvement on the Zebra problem, $n$-Queens problems, and random binary problems. Bayardo and Schrag (1996, 1997) show that adding CBJ to the well-known Davis-Putnam algorithm, the SAT version of forward checking, can be a significant improvement on hard random and real-world 3-SAT problems.

Adding CBJ to an algorithm that maintains full arc consistency has received less attention in the literature. In the one study that we are aware of, Bessière and Régin (1996) observe that adding CBJ to MAC (the binary version of GAC) actually slows down the algorithm on random binary problems due to the overhead of maintaining the conflict sets. They conjecture that "when MAC and a good variable ordering heuristic are used, CBJ becomes useless".

Our empirical results lead us to differ with Bessière and Régin's conclusion about the usefulness of adding CBJ to an algorithm that maintains full arc consistency. In our implementation we were able to significantly reduce the overhead of maintaining the conflict sets through the use of additional data structures[5]. On problems where adding CBJ does not lead to many savings in nodes visited, our implementation of CBJ also does not degrade performance by any significant factor. We demonstrate the improvement by re-doing Bessière and Régin's (1996) experiments on random binary problems. We then show through experiments in two structured domains that GAC-CBJ can sometimes improve GAC by several orders of magnitude on hard instances.

In our experiments, we ran both GAC and GAC-CBJ on each instance of a problem and recorded the CPU times. Comparing CPU times is appropriate as the underlying code for GAC and GAC-CBJ is identical, with GAC-CBJ containing only *additional* code to maintain the conflict sets and to determine how far to jump back. Two dynamic variable orderings were used: the popular *dom+deg* heuristic which chooses the next variable with the minimal domain size and breaks ties by choosing the variable with the maximum degree (the number of the constraints that constrain that variable) and the *dom/deg* heuristic proposed by Bessière and Régin (1996) which chooses the next variable with the minimal

5. See the online appendix for the source code and a description of the key data structures in our implementations of GAC and GAC-CBJ.

value of the domain size divided by its degree. All experiments were run on 400 MHz Pentium II's with 256 Megabytes of memory.

### 5.1 Random Problems

The run time performance of GAC and GAC-CBJ were compared on sets of randomly generated binary CSPs. A set of random problems is defined by a 5-tuple $(n, d, r, m, t)$, where $n$ is the number of the variables, $d$ is the uniform domain size, $r$ is the uniform arity of the constraints, $m$ is the number of randomly generated constraints, and $t$ is the uniform tightness or number of tuples in each constraint. In each case, the constraint tightness $t$ was chosen so that approximately half of the instances in the population were insoluble; i.e., the instances were from the phase transition region.

Table 1: Effect of domain size on average time (seconds) to solve random instances from $(50, d, 2, 95, t)$. Each set contained 100 random instances. Both GAC-CBJ and GAC used the *dom/deg* variable ordering.

| $d$ | GAC-CBJ | GAC | ratio |
|---|---|---|---|
| 5 | 0.0027 | 0.0030 | 0.90 |
| 10 | 0.026 | 0.027 | 0.96 |
| 15 | 0.10 | 0.10 | 1.00 |
| 20 | 0.41 | 0.41 | 1.00 |
| 25 | 0.79 | 0.78 | 1.01 |
| 30 | 2.46 | 2.47 | 1.00 |
| 35 | 3.82 | 3.80 | 1.01 |
| 40 | 10.98 | 10.75 | 1.02 |

Bessière and Règin (1996) examine the effect of domain size on the average time to solve random instances from $(50, d, 2, 95, t)$ (see Figure 5 (right) in Bessière & Régin, 1996). With their implementation of CBJ, adding CBJ steadily worsens performance as domain size increases until at $d = 40$ MAC-CBJ is about 1.7 times slower than MAC alone. With our implementation, the difference in performance between GAC-CBJ and GAC was negligible on these problems (see Table 1).

The remaining sets of random problems that Bessière and Règin used in their experiments to compare the performance of MAC-CBJ and MAC are now too simple to provide a meaningful comparison as they can be solved in less than 0.01 seconds on a 400 MHz Pentium II computer. Thus, we chose harder sets of random binary problems. On each instance we ran both GAC and GAC-CBJ and recorded the CPU times. Here we report the average ratio of the CPU times (GAC over GAC-CBJ). Each set contained 100 random instances. On the first set of problems, $(150, 5, 2, 750, 19)$, the average ratio for the *dom+deg* variable ordering was 0.90 and the average ratio for the *dom/deg* variable ordering was 0.88. On the second set of problems, $(150, 5, 2, 1500, 21)$, the average ratios for both the *dom+deg* and *dom/deg* variable orderings was 0.93. In other words, on average GAC was a little over 10% faster than GAC-CBJ on these problems.

### 5.2 Planning Problems

Planning, where one is required to find a sequence of actions from an initial state to a goal state, can be formulated as a CSP. In the formulation we used in our experiments, each state is modeled by a collection of variables and the constraints enforce the assignments of variables to represent a consistent state or a valid transition between states. (See Kautz & Selman, 1992; van Beek & Chen, 1999 for more details on the formulation of planning as a CSP.)

Table 2: Time (seconds) to solve instances of the grid planning problem. The absence of an entry indicates that the problem was not solved within 72000 seconds (20 hours) of CPU time.

| | *dom+deg* | | *dom/deg* | |
|---|---|---|---|---|
| | GAC | GAC-CBJ | GAC | GAC-CBJ |
| 1 | 0.66 | 0.68 | 1.58 | 0.86 |
| 2 | **762.47** | **33.33** | **3965.10** | **321.17** |
| 3 | . | . | . | . |
| 4 | . | **1753.13** | . | . |
| 5 | . | . | . | . |

In the experiments we used all 130 instances used in the First AI Planning Systems Competition, June 6–9, 1998. The instances come from five different domains: gripper, mystery, mprime, logistics, and grid. In the experiments we report, both GAC and GAC-CBJ were based on AC3 (Mackworth, 1977a) as this was found to give the best performance.

For the gripper, mystery, and mprime domains, each of the instances could be solved in under 25 seconds by both GAC and GAC-CBJ. On these easy problems, the increased overhead of CBJ rarely led to savings, and overall GAC was 10-15% faster than GAC-CBJ.

Table 2 shows the comparison between GAC and GAC-CBJ in solving the 5 instances of the grid problems. GAC-CBJ showed improvement on the grid problems. For example, it solved problem 4 in about half an hour, but GAC failed to find a solution in 20 hours.

Table 3 shows the comparison between GAC and GAC-CBJ in solving the 30 instances of the logistics problem. On about one third of the instances, GAC-CBJ improved on GAC. For example, on instances 18, 20 and 27, GAC-CBJ ran several orders of magnitude faster than GAC, and on instance 15, GAC exhausted the 20 hours time limit but GAC-CBJ found a solution within 3 minutes. GAC-CBJ and GAC performed similarly on easier instances and sometimes GAC-CBJ was about 10% slower than GAC.

Table 3: Time (seconds) to solve instances of the logistics planning problem. The absence of an entry indicates that the problem was not solved within 72000 seconds (20 hours) of CPU time.

| | *dom+deg* | | *dom/deg* | |
|---|---|---|---|---|
| | GAC | GAC-CBJ | GAC | GAC-CBJ |
| 1 | 0.03 | 0.03 | 0.03 | 0.03 |
| 2 | 0.03 | 0.05 | 0.03 | 0.06 |
| 3 | **10.91** | **0.86** | **9.63** | **0.81** |
| 4 | 0.16 | 0.17 | 0.14 | 0.18 |
| 5 | 1.51 | 1.54 | 1.54 | 1.57 |
| 6 | **36.49** | **16.86** | **35.77** | **16.76** |
| 7 | 0.08 | 0.08 | 0.08 | 0.09 |
| 8 | 0.15 | 0.15 | 0.14 | 0.16 |
| 9 | 0.30 | 0.33 | 0.32 | 0.33 |
| 10 | . | . | . | . |
| 11 | 0.04 | 0.05 | 0.05 | 0.05 |
| 12 | 0.11 | 0.13 | 0.11 | 0.11 |
| 13 | 0.54 | 0.57 | 0.54 | 0.56 |
| 14 | 0.63 | 0.64 | 0.64 | 0.68 |
| 15 | **.** | **182.51** | **.** | **8540.58** |
| 16 | **12.49** | **0.42** | **12.32** | **0.41** |
| 17 | **264.46** | **0.32** | **261.33** | **0.32** |
| 18 | **15382.82** | **1165.54** | **15157.71** | **1184.67** |
| 19 | 1.29 | 1.37 | 1.33 | 1.31 |
| 20 | **6268.16** | **27.66** | **6125.87** | **28.55** |
| 21 | 0.66 | 0.70 | 0.68 | 0.74 |
| 22 | . | . | . | . |
| 23 | . | . | . | . |
| 24 | 0.08 | 0.09 | 0.08 | 0.09 |
| 25 | 34.03 | 13.03 | 11.58 | 12.10 |
| 26 | . | . | . | . |
| 27 | **12239.26** | **47.06** | **12105.62** | **47.76** |
| 28 | . | . | . | . |
| 29 | . | . | . | . |
| 30 | . | . | . | . |

### 5.3 Crossword Puzzle Problems

Crossword puzzle generation, where one is required to fill in a grid with words from a dictionary, can be formulated as a CSP. In the formulation we used in our experiments, each of the unknown words is represented by a variable which takes values from the dictionary. Binary constraints enforce that intersecting words agree on their intersecting letter and that a word from the dictionary appears at most once in a solution. Figure 7 shows an example $5 \times 5$ crossword puzzle grid. A CSP model of this grid has 10 variables, 21 binary "intersection" constraints, and 13 "not equals" constraints.

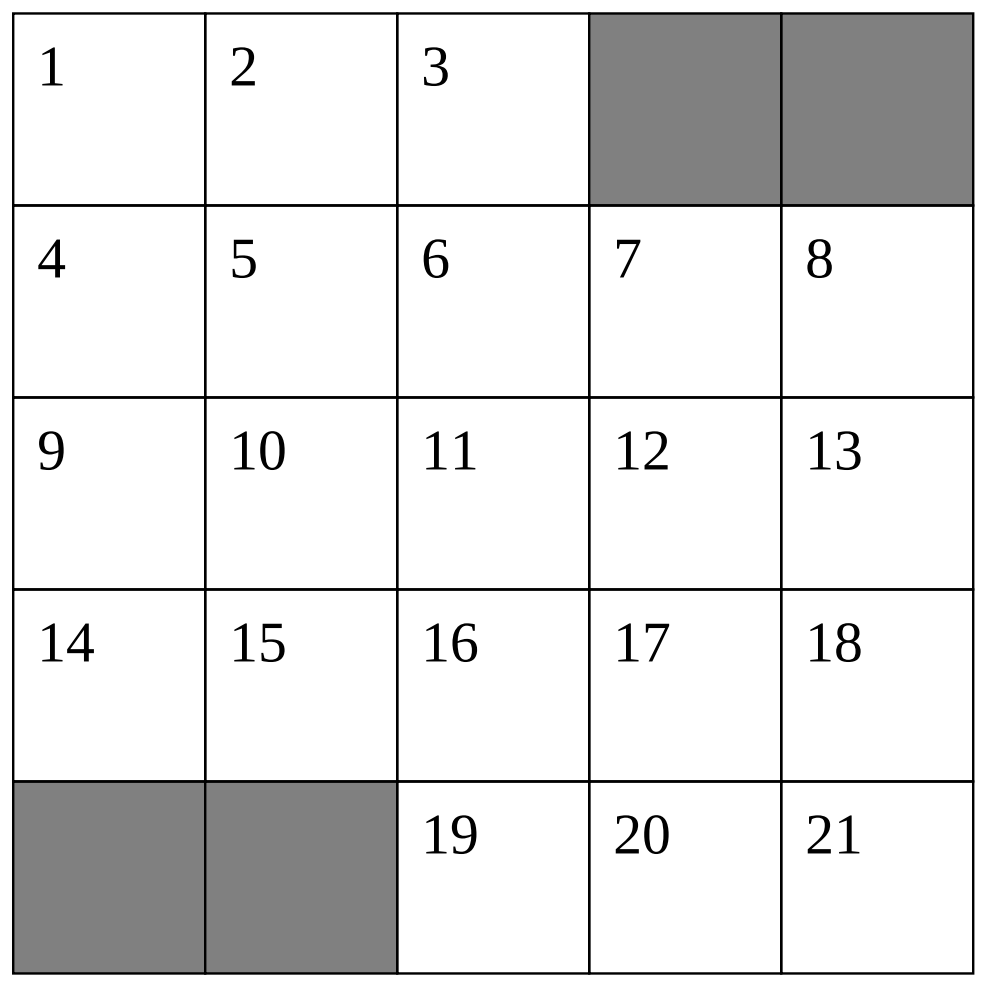

Figure 7: A crossword puzzle.

In the experiments we used 50 grids and two dictionaries for a total of 100 instances of the problem that ranged from easy to very hard. For the grids, we used 10 instances at each of the following sizes: $5{\times}5$, $15{\times}15$, $19{\times}19$, $21{\times}21$, and $23{\times}23$. For the dictionaries we used the UK dictionary, which collects about 220,000 words and in which the largest domain for a word variable contains about 30,000 values, and the Linux dictionary, which collects 45,000 words and in which the largest domain for a word variable has about 5,000 values. In the experiments we report, both GAC and GAC-CBJ were based on AC7 (Bessière & Régin, 1997) as this was found to give the best performance (see Sillito, 2000 for a discussion of integrating AC7 into backtracking search).

Figure 8 shows approximate cumulative frequency curves for the empirical results, where we are plotting the ratio of the time taken to solve an instance by GAC over the time taken to solve the instance by GAC-CBJ. Thus, for example, we can read from the curve representing the *dom+deg* variable ordering that for approximately 85% of the tests adding CBJ had little effect and that for the remaining 15% of the tests it led to orders of magnitude improvements. We can also read from the curves the 0, 10, ..., 100 percentiles of the data sets (where the value of the median is the 50*th* percentile or the value of the 50*th* test). The crossover point, where GAC-CBJ starts to perform as well as or better than GAC occurs around the 35*th* percentile. Tables 4 and 5 examine the data more closely by showing the actual times to solve the instances where GAC performed best and the instances where GAC-CBJ performed best.

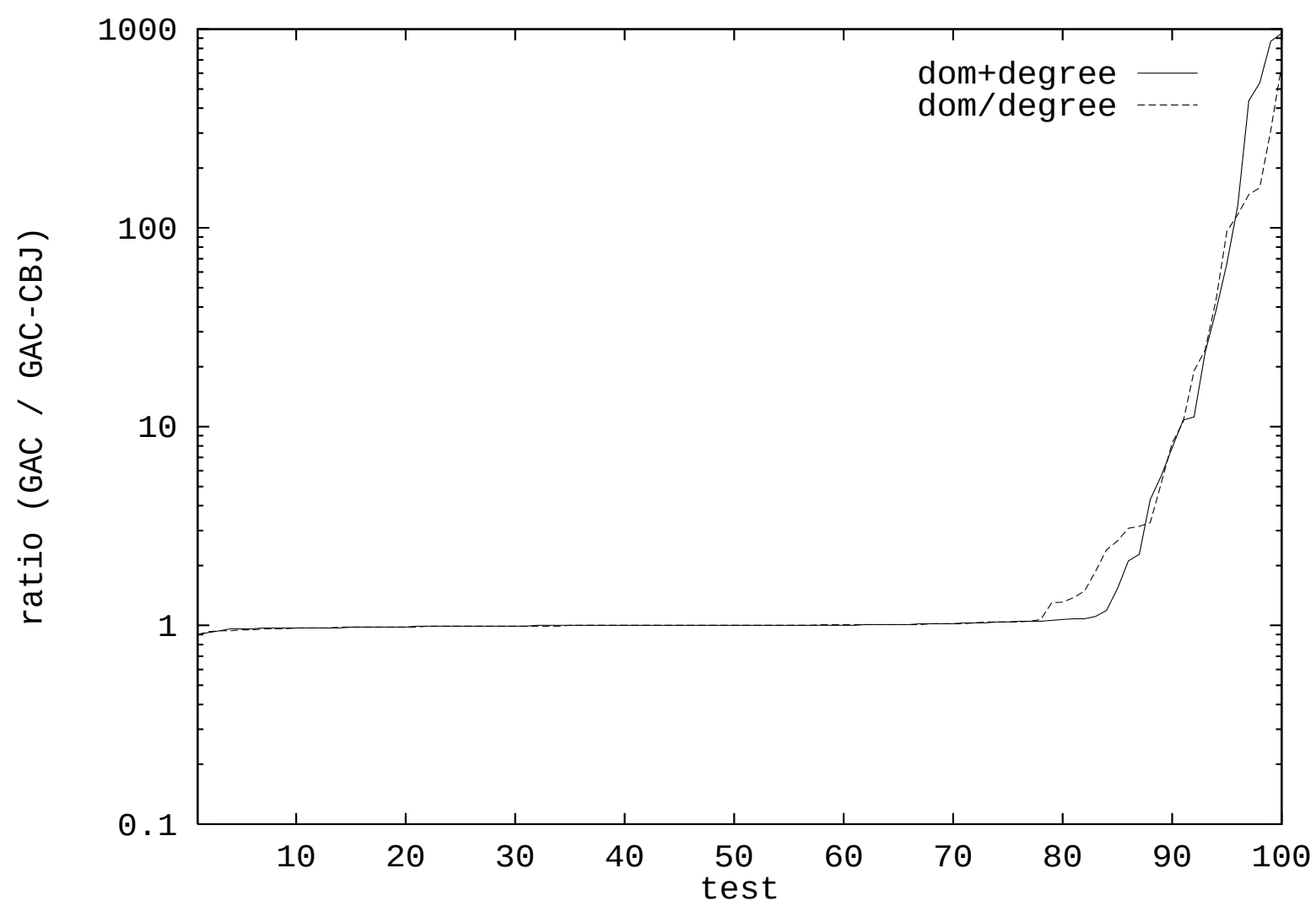

Figure 8: Effect on execution time of GAC of adding conflict-directed backjumping (GAC-CBJ). A curve represents 100 tests on instances of the crossword puzzle problem where the tests are ordered by the ratio of time taken to solve the instance by GAC over time taken to solve the instance by GAC-CBJ.

Table 4: GAC versus GAC-CBJ on instances of the crossword puzzle problem. The ten best improvements in time (seconds) of GAC over GAC-CBJ to solve an instance are presented.

| | *dom+deg* | | *dom/deg* | |
|---|---|---|---|---|
| rank | GAC | GAC-CBJ | GAC | GAC-CBJ |
| 1 | 1.21 | 1.35 | 1.11 | 1.23 |
| 2 | 1.10 | 1.20 | 0.95 | 1.02 |
| 3 | 6.12 | 6.53 | 1.16 | 1.24 |
| 4 | 0.78 | 0.81 | 56.66 | 60.36 |
| 5 | 110.23 | 114.52 | 1.30 | 1.37 |
| 6 | 68.67 | 71.28 | 4.86 | 5.11 |
| 7 | 47.16 | 48.42 | 0.22 | 0.23 |
| 8 | 32.69 | 33.63 | 14.23 | 14.76 |
| 9 | 25.17 | 26.08 | 74.38 | 77.52 |
| 10 | 20.73 | 21.37 | 7.43 | 7.67 |

Table 5: GAC versus GAC-CBJ on instances of the crossword puzzle problem. The ten best improvements in time (seconds) of GAC-CBJ over GAC to solve an instance are presented. The absence of an entry indicates that the problem was not solved within 36000 seconds (10 hours) of CPU time.

| | *dom+deg* | | *dom/deg* | |
|---|---|---|---|---|
| rank | GAC | GAC-CBJ | GAC | GAC-CBJ |
| 1 | . | 37.85 | . | 54.60 |
| 2 | . | 41.43 | 10311.32 | 33.43 |
| 3 | . | 67.07 | . | 225.92 |
| 4 | . | 82.58 | . | 244.81 |
| 5 | . | 276.00 | . | 308.04 |
| 6 | . | 542.80 | . | 374.72 |
| 7 | . | 939.71 | . | 832.68 |
| 8 | 2716.86 | 115.87 | . | 1486.43 |
| 9 | 390.91 | 34.90 | . | 1890.24 |
| 10 | . | 3336.37 | . | 3411.83 |

In summary, on some of the smaller, easier crossword puzzle instances GAC was slightly faster than GAC-CBJ, on many of the puzzles there was no noticeable difference, and on some of the larger, harder puzzles GAC-CBJ was orders of magnitude faster than GAC.

## 6. Conclusion

In this paper, we presented three main results. First, we showed that the choice of dynamic variable ordering heuristic can weaken the effects of the backjumping technique. Second, we showed that as the level of local consistency that is maintained in the backtracking search is increased, the less that backjumping will be an improvement. Together these results partially explain why a backtracking algorithm doing more in the look-ahead phase cannot benefit more from the backjumping look-back scheme and they extend the partial ordering of backtracking algorithms presented by Kondrak and van Beek (1997) to include backtracking algorithms and their CBJ hybrids that maintain levels of local consistency beyond forward checking. Third, and finally, we showed that adding CBJ to a backtracking algorithm that maintains generalized arc consistency can (still) speed up the algorithm by several orders of magnitude on hard, structured problems. Throughout the paper, we did not restrict ourselves to binary CSPs.

## Acknowledgements

The authors wish to thank the referees for their careful reading of a previous version of the paper and their helpful comments. The financial support of the Canadian Government through their NSERC program is gratefully acknowledged.

## References

Bacchus, F., & Grove, A. (1999). Looking forward in constraint satisfaction algorithms. Unpublished manuscript.

Bacchus, F., & van Run, P. (1995). Dynamic variable ordering in CSPs. In *Proceedings of the First International Conference on Principles and Practice of Constraint Programming*, pp. 258–275, Cassis, France. Available as: Springer Lecture Notes in Computer Science 976.

Bayardo Jr., R. J., & Schrag, R. (1996). Using CSP look-back techniques to solve exceptionally hard SAT instances. In *Proceedings of the Second International Conference on Principles and Practice of Constraint Programming*, pp. 46–60, Cambridge, Mass. Available as: Springer Lecture Notes in Computer Science 1118.

Bayardo Jr, R. J., & Schrag, R. C. (1997). Using CSP look-back techniques to solve real-world SAT instances. In *Proceedings of the Fourteenth National Conference on Artificial Intelligence*, pp. 203–208, Providence, RI.

Bessière, C., & Régin, J.-C. (1996). MAC and combined heuristics: Two reasons to forsake FC (and CBJ?) on hard problems. In *Proceedings of the Second International Conference on Principles and Practice of Constraint Programming*, pp. 61–75, Cambridge, Mass.

Bessière, C., & Régin, J.-C. (1997). Arc consistency for general constraint networks: Preliminary results. In *Proceedings of the Sixteenth International Joint Conference on Artificial Intelligence*, pp. 398–404, Nagoya, Japan.

Bruynooghe, M. (1981). Solving combinatorial search problems by intelligent backtracking. *Information Processing Letters*, *12*, 36–39.

Chen, X. (2000). *A Theoretical Comparison of Selected CSP Solving and Modeling Techniques*. Ph.D. thesis, University of Alberta.

Dechter, R. (1990). Enhancement schemes for constraint processing: Backjumping, learning, and cutset decomposition. *Artificial Intelligence*, *41*, 273–312.

Dechter, R. (1992). Constraint networks. In Shapiro, S. C. (Ed.), *Encyclopedia of Artificial Intelligence, 2nd Edition*, pp. 276–285. John Wiley & Sons.

Freuder, E. C. (1978). Synthesizing constraint expressions. *Comm. ACM*, *21*, 958–966.

Frost, D., & Dechter, R. (1994). Dead-end driven learning. In *Proceedings of the Twelfth National Conference on Artificial Intelligence*, pp. 294–300, Seattle, Wash.

Gaschnig, J. (1978). Experimental case studies of backtrack vs. Waltz-type vs. new algorithms for satisficing assignment problems. In *Proceedings of the Second Canadian Conference on Artificial Intelligence*, pp. 268–277, Toronto, Ont.

Haralick, R. M., & Elliott, G. L. (1980). Increasing tree search efficiency for constraint satisfaction problems. *Artificial Intelligence*, *14*, 263–313.

Kautz, H., & Selman, B. (1992). Planning as satisfiability. In *Proceedings of the 10th European Conference on Artificial Intelligence*, pp. 359–363, Vienna.

Kondrak, G., & van Beek, P. (1997). A theoretical evaluation of selected backtracking algorithms. *Artificial Intelligence*, *89*, 365–387.

Mackworth, A. K. (1977a). Consistency in networks of relations. *Artificial Intelligence*, *8*, 99–118.

Mackworth, A. K. (1977b). On reading sketch maps. In *Proceedings of the Fifth International Joint Conference on Artificial Intelligence*, pp. 598–606, Cambridge, Mass.

McGregor, J. J. (1979). Relational consistency algorithms and their application in finding subgraph and graph isomorphisms. *Inform. Sci.*, *19*, 229–250.

Montanari, U. (1974). Networks of constraints: Fundamental properties and applications to picture processing. *Inform. Sci.*, *7*, 95–132.

Nadel, B. A. (1989). Constraint satisfaction algorithms. *Computational Intelligence*, *5*, 188–224.

Prosser, P. (1993a). Domain filtering can degrade intelligent backtracking search. In *Proceedings of the Thirteenth International Joint Conference on Artificial Intelligence*, pp. 262–267, Chambèry, France.

Prosser, P. (1993b). Hybrid algorithms for the constraint satisfaction problem. *Computational Intelligence*, *9*, 268–299.

Prosser, P. (1995). MAC-CBJ: Maintaining arc consistency with conflict-directed backjumping. Research report 177, University of Strathclyde.

Sabin, D., & Freuder, E. C. (1994). Contradicting conventional wisdom in constraint satisfaction. In *Proceedings of the 11th European Conference on Artificial Intelligence*, pp. 125–129, Amsterdam.

Schiex, T., & Verfaillie, G. (1994). Nogood recording for static and dynamic constraint satisfaction problems. *International Journal on Artificial Intelligence Tools*, *3*, 1–15.

Sillito, J. (2000). Improving and Estimating the Cost of Backtracking Algorithms for CSPs.. MSc thesis, University of Alberta, 2000.

Smith, B. M., & Grant, S. A. (1995). Sparse constraint graphs and exceptionally hard problems. In *Proceedings of the Fourteenth International Joint Conference on Artificial Intelligence*, pp. 646–651, Montreal.

van Beek, P., & Chen, X. (1999). CPlan: A constraint programming approach to planning. In *Proceedings of the Sixteenth National Conference on Artificial Intelligence*, pp. 585–590, Orlando, Florida.